\documentclass{article}
\usepackage[preprint, nonatbib]{neurips_2023}
\pdfoutput=1

\usepackage[utf8]{inputenc} 
\usepackage[T1]{fontenc}    
\usepackage{url}            
\usepackage{booktabs}       
\usepackage{amsfonts}       
\usepackage{nicefrac}       
\usepackage{microtype}      
\usepackage{xcolor}         
\usepackage{bbding}         
\usepackage{graphicx}
\usepackage{tabularx}       
\usepackage{comment}

\usepackage{amsmath}
\usepackage{amssymb}
\usepackage{float}
\usepackage[symbol]{footmisc}
\usepackage{multirow}
\usepackage{xcolor}
\usepackage[colorlinks = true,
            urlcolor  = red,
            ]{hyperref}
            
\usepackage[capitalize]{cleveref}
\crefname{section}{Sec.}{Secs.}
\Crefname{section}{Section}{Sections}
\Crefname{table}{Table}{Tables}
\crefname{table}{Tab.}{Tabs.}

\usepackage{hyphenat}
\usepackage{soul,color}
\usepackage{amsopn}

\definecolor{Red}{cmyk}{0,1,1,0}
\definecolor{Green}{cmyk}{1,0,1,0}
\definecolor{Cyan}{cmyk}{1,0,0,0}
\definecolor{Purple}{cmyk}{0.45,0.86,0,0}
\definecolor{Rosolic}{cmyk}{0.00,1.00,0.50,0}
\definecolor{Blue}{cmyk}{1.00,1.00,0.00,0}
\definecolor{BlueViolet}{cmyk}{0.86,0.91,0,0.04}
\definecolor{NavyBlue}{cmyk}{0.94,0.54,0,0}

\newcommand{\cx}[1]{{\color{NavyBlue} CX:#1}}

\newcommand{\hidden}[1]{{\color{NavyBlue}}}
\newcommand{\ct}[1]{{#1}}

\title{MotionGPT: Human Motion as a Foreign Language}

\author{Biao Jiang$^{1,2}$\thanks{Contributed equally and work done while Biao Jiang was a Research Intern with Tencent PCG.} \quad\quad Xin Chen$^{2}$\footnotemark[1] \quad\quad Wen Liu$^{2}$ \quad\quad Jingyi Yu$^{3}$ \quad\quad Gang Yu$^{2}$ \quad\quad Tao Chen$^{1}$\thanks{Corresponding author.} 
\\ $^{1}$Fudan University \quad\quad\quad$^{2}$Tencent PCG \quad\quad\quad$^{3}$ ShanghaiTech University \\
\tt \small \textbf{\href{https://github.com/OpenMotionLab/MotionGPT}{https://github.com/OpenMotionLab/MotionGPT}}
}

\begin{document}

\maketitle

\begin{abstract}
  Though the advancement of pre-trained large language models unfolds, the \ct{exploration} of \ct{building} a unified model for language and other multimodal data, such as motion, \ct{remains challenging and untouched so far.} 
  %
  %
  Fortunately, human motion displays a semantic coupling akin to human language, often perceived as a form of body language.
  By fusing language data with \ct{large-scale motion} models,
  motion-language pre-training that can enhance the performance of motion-related tasks becomes feasible. 
  %
  Driven by this insight, we propose MotionGPT, a unified, \ct{versatile, and user-friendly} motion-language model to handle \ct{multiple} motion-relevant tasks.
  %
  Specifically, we employ the discrete vector quantization for human motion and transfer 3D motion into motion tokens, similar to \ct{the generation process of} word tokens. 
  Building upon this ``motion vocabulary'', we perform language modeling on both motion and text in a unified manner, treating human motion as a specific language.
  Moreover, inspired by prompt learning, we pre-train MotionGPT with a mixture of motion-language data and fine-tune it on prompt-based question-and-answer tasks.
  %
  Extensive experiments demonstrate that MotionGPT achieves state-of-the-art performances on multiple motion tasks including text-driven motion generation, motion captioning, motion prediction, and motion in-between.
  
\end{abstract}

\section{Introduction}
\label{intro}

Recent years have witnessed a significant breakthrough in pre-trained large language models such as GPT~\cite{radford2018gpt,radford2019gpt2,brown2020gpt3,ouyang2022instructgpt}, BERT~\cite{devlin2018bert}, and T5~\cite{raffel2020t5,chung2022flant5}, which lead to the convergence of language~\cite{zhang2022opt,touvron2023llama}, image~\cite{radford2021clip,wang2022beit,li2022blip}, mesh~\cite{youwang2022clipactor,mohammad2022clipmesh} and mutlimodal~\cite{girdhar2023imagebind} modeling. Nevertheless, a general pre-trained model for human motion and language has yet to emerge. This pre-trained motion-language model, capable of supporting numerous motion-relevant tasks through prompts, should benefit diverse fields like gaming, robotics, virtual assistant, and human behavior analysis.

%
Previous research on human motion has explored various tasks, including motion generation~\cite{petrovich21actor, Guo_2022_CVPR_humanml3d,mdm2022human,chen2023mld,zhang2023generating}, motion captioning~\cite{goutsu2021seqgan,chuan2022tm2t}, and motion prediction~\cite{yuan2020dlow,zhang2021we,ma2022multi}. 
Recent text-to-motion works\cite{mdm2022human, zhang2022motiondiffuse,petrovich22temos,chen2023mld} have attempted to employ pre-trained language-relevant models~\cite{devlin2018bert,radford2021clip}. For instance, MDM~\cite{mdm2022human} learns a motion diffusion model with conditional text tokens from CLIP~\cite{radford2021clip}, while MLD~\cite{chen2023mld} integrates motion latent space to improve the efficiency of motion diffusion process.
On the other hand, MotionCLIP~\cite{tevet2022motionclip} and TM2T~\cite{chuan2022tm2t} concentrate on modeling the coupled relationship between motion and text description.
However, the above approaches treat motion and language as separate modalities, which often require \ct{strictly} paired motion and text data.
Moreover, since the supervisions are task-specific, they can hardly generalize effectively to unseen tasks or data, as they lack a comprehensive understanding of the relationship between motion and language.
We thus focus on \ct{building} a pre-trained motion-language model, which can generalize to various tasks and learn \ct{in-depth motion-language correlation knowledge} from more feasible motion and language data.

\begin{figure}[t] 
	\centering 
	\includegraphics[width=\linewidth]{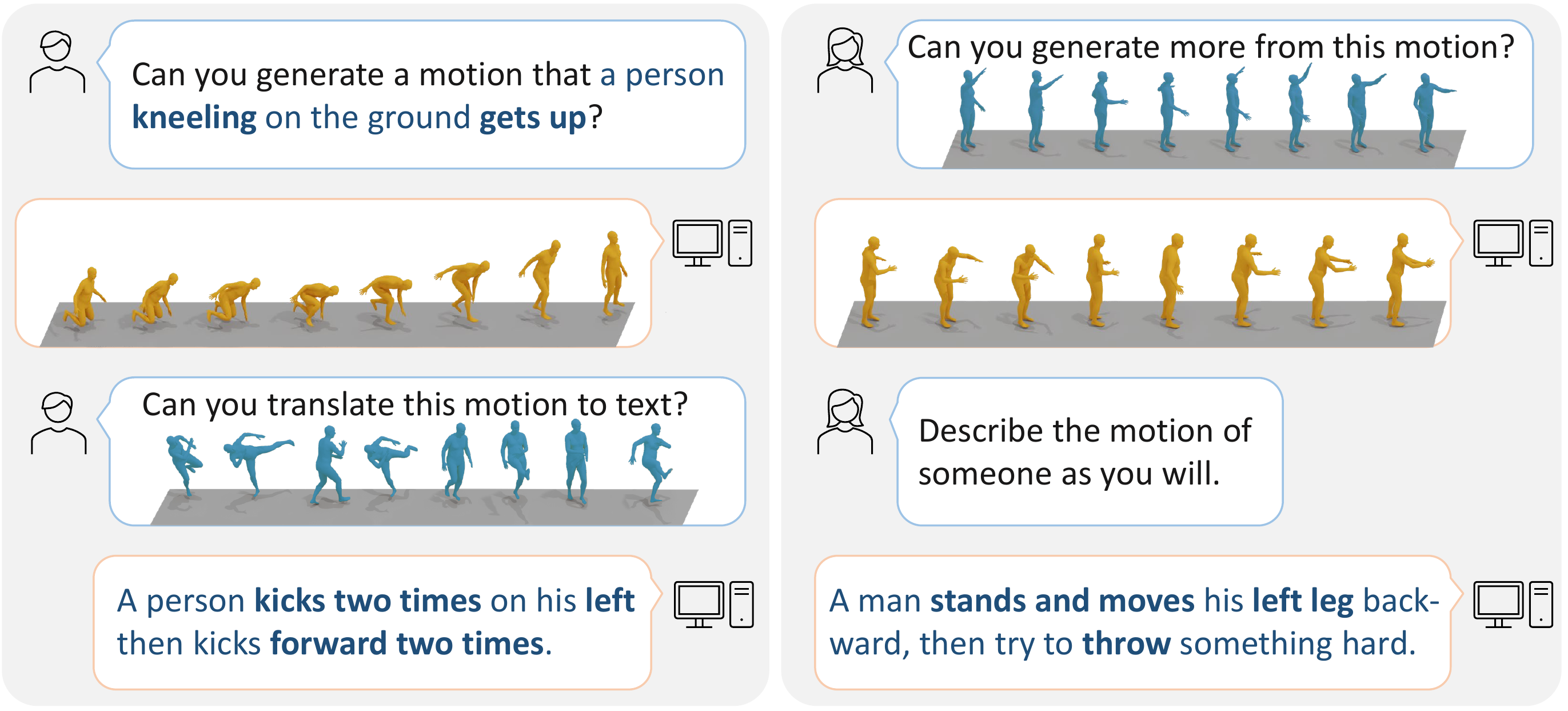} 
 \vspace{-18pt}
\caption{MotionGPT can address diverse motion-relevant tasks uniformly given different instructions. We provide the results on text-to-motion (the upper left), motion captioning (the bottom left), motion completion (the upper right), and the language question-to-answer (the bottom right). The left to right of motion represents the time order. Blue motion denotes the input, and yellow is the generation.} 
\vspace{-10pt}
	\label{fig:teaser} 
\end{figure}

Two challenges are crucial \ct{and need to be solved} for \ct{pre-training a promising motion-language model}. The first is modeling the relation between language and motion, and the second is \ct{building} a uniform \ct{multi-task framework} that can generalize to new tasks.
%
Fortunately, human motion exhibits a semantic coupling similar to human language, often interpreted as a form of body language.
Building upon this observation,
we \ct{follow vision-language pre-training from BEiT-3~\cite{wang2022beit} to
} treat human motion as a specific foreign language. 
By integrating motion and language data \ct{together and encoding them within} a single vocabulary, the relationship between motion and language 
becomes more apparent.
Therefore, with recent significantly larger-scale language data and models, the motion-language pre-training \ct{has great potential to} improve the performance on motion tasks.
Meanwhile, this pre-training on language enables textual instructions like prompts in InstructGPT~\cite{ouyang2022instructgpt} and makes the model more versatile and user-friendly for various motion tasks.

In this work, we propose a uniform motion-language framework, namely MotionGPT, that leverages the strong language generation and zero-shot transfer abilities of pre-trained language models for doing human motion-related tasks. 
To enable MotionGPT to comprehend and generate human-like motions, we first learn a motion-specific vector quantized variational autoencoder (VQ-VAE) model to construct ``motion vocabulary'', akin to English vocabulary and then convert raw motion data into a sequence of motion tokens.
These tokens are then processed by a pre-trained language model~\cite{raffel2020t5,chung2022flant5} that learns the underlying grammar and syntax of the motion language, as well as its relationship with the corresponding textual descriptions.
To effectively integrate language and motion in MotionGPT, we design a two-stage training scheme.
We first pre-train the language model on the raw motion dataset to learn the basic grammar and syntax of the motion language.
For prompt tuning, we fine-tune the language model on an instruction dataset, which contains both textual descriptions and motion data, to learn the correlation between the two modalities.
%
Extensive experiments demonstrate that MotionGPT achieves state-of-the-art performance on text-to-motion, motion-to-text, motion prediction, and motion in-between.
%

We summarize our contributions as follows: 
(1) We propose a uniform motion-language generative pre-trained model, MotionGPT, which treats human motion as a foreign language, introduces natural language models into motion-relevant generation, and performs diverse motion tasks with a single model.
(2) We introduce a motion-language training scheme with instruction tuning, to learn from task feedback and produce promising results through prompts.
(3) We propose a general motion benchmark for multi-task evaluation, wherein MotionGPT achieves competitive performance across diverse tasks, including text-to-motion, motion-to-text, motion prediction, and motion in-between, with all available codes and data.

\section{Related Work}
\label{relatedwork}

\textbf{Human Motion Synthesis} involves generating diverse and realistic human-like motion using multimodal inputs, such as text~\cite{Guo_2022_CVPR_humanml3d,petrovich22temos,zhang2022motiondiffuse,mdm2022human,chuan2022tm2t,ahuja2019language2pose,kim2022flame}, action~\cite{petrovich21actor, guo2020action2motion,mdm2022human,chen2023mld}, and incomplete motion ~\cite{yuan2020dlow, zhang2021we, ma2022multi,mdm2022human}.
Text-to-motion is one of the most important motion generation tasks, due to the user-friendly and convenient language input. MDM~\cite{mdm2022human} proposes a diffusion-based generative model~\cite{ho2020denoising} separately trained on several motion tasks. MLD~\cite{chen2023mld} advances the latent diffusion model~\cite{song2020denoising,stable_diffusion} to generate motions based on different conditional inputs.
T2M-GPT~\cite{zhang2023generating} investigates a generative framework based on VQ-VAE and Generative Pre-trained Transformer (GPT) for motion generation. 
Motion completion task generates motion conditioning on partial motions, such as classical motion prediction \cite{yuan2020dlow, zhang2021we, ma2022multi} or motion in-between \cite{mdm2022human}, which generates the intermediate motion while the first and last parts are fixed.
Although they show promising results in various human motion tasks, most above methods are limited in using a single model to handle multiple tasks. We thus propose a uniform approach that treats human motion as a foreign language, and leverages the strong language generation and zero-shot transfer abilities of pre-trained language models 

\vspace{20pt}
\begin{table}[t]
\resizebox{\columnwidth}{!}{%
\begin{tabular}{@{}lcccccccccc@{}}
\toprule
Methods & Text-to-Motion & Motion-to-Text & Motion Prediction & Motion In-between & Random Motion & Random Description
\\ \midrule
T2M-GPT \cite{mdm2022human}& \CheckmarkBold & \XSolidBrush & \XSolidBrush & \XSolidBrush & \CheckmarkBold & \XSolidBrush  \\
MLD \cite{chen2023mld} & \CheckmarkBold & \XSolidBrush & \XSolidBrush & \XSolidBrush& \CheckmarkBold & \XSolidBrush 
  \\
TM2T \cite{chuan2022tm2t} & \CheckmarkBold & \CheckmarkBold  & \XSolidBrush & \XSolidBrush & \XSolidBrush & \XSolidBrush
  \\
MDM \cite{mdm2022human}   & \CheckmarkBold   & \XSolidBrush  & \CheckmarkBold & \CheckmarkBold & \CheckmarkBold & \XSolidBrush \\
MotionDiffuse\cite{zhang2022motiondiffuse}  & \CheckmarkBold   & \XSolidBrush  & \CheckmarkBold & \CheckmarkBold & \CheckmarkBold & \XSolidBrush 
  \\
 \midrule
MotionGPT (Ours) & \CheckmarkBold& \CheckmarkBold& \CheckmarkBold& \CheckmarkBold& \CheckmarkBold& \CheckmarkBold
\\\bottomrule
\end{tabular}%
}
\label{tab:checkmark}
\vspace{5pt}
\caption{Comparison of recent state-of-the-art methods on diverse motion-relevant tasks. \textit{Random Motion} and \textit{Random Caption} represent unconstrained generation of motions and motion descriptions.}
\vspace{-10pt}
\end{table}

\textbf{Human Motion Captioning.} To describe human motion with natural languages, \cite{takano2015statistical} learns the mapping from motions to language relying on two statistical models.
Furthermore, recurrent networks have also been used in \cite{yamada2018rae,plappert2018learning}. 
More recently, TM2T~\cite{chuan2022tm2t} proposed a new motion representation that compresses motions into a short sequence of discrete variables, then uses a neural translation network to build mappings between two modalities. While previous research like TM2T~\cite{chuan2022tm2t} incorporated captioning modules into their training pipeline for motion generation, these approaches are constrained to bidirectional translation between text and motion within one uniform framework.

\textbf{Language Models and Multi-Modal.}
Large-scale language models (LLMs)~\cite{devlin2018bert,dai2019transformer,raffel2020t5,brown2020gpt3,zhang2022opt,touvron2023llama}, enabled by extensive datasets and model size, have demonstrated impressive comprehension and generation capabilities, elevating natural language processing to new heights. BERT~\cite{devlin2018bert} pre-trains deep bidirectional language representations that can support downstream tasks. T5~\cite{raffel2020t5} introduced a unified framework that converts all text-based language problems into a text-to-text format. More recent research~\cite{wei2021finetuned,bach2022promptsource,ouyang2022instructgpt,chung2022flant5} find that by fine-tuning pre-trained models using input-output pairs consisting of instructions and coupled answers, the performance of pre-trained models can be further improved. FLAN~\cite{chung2022flant5} presents an instruction-tuning technique that surpasses the performance of non-tuned models in unseen tasks. 
Recently, the wave of multi-modal models~\cite{li2022blip,huang2023language,li2023blip2} is intriguing to process text along with other modalities, such as images~\cite{li2022blip,huang2023language,girdhar2023imagebind}, audio~\cite{guzhov2022audioclip,girdhar2023imagebind}, and videos~\cite{xu2021videoclip}. CLIP~\cite{radford2021clip} further learns a semantic latent representation that couples images with corresponding language descriptions.
Despite the success of language models in various vision-language tasks, the development of multi-modal language models that can handle human motion is still limited.

\textbf{Motion Language Pre-training.} Existing text-to-motion generation methods~\cite{Guo_2022_CVPR_humanml3d,petrovich22temos, mdm2022human,chuan2022tm2t,ahuja2019language2pose,kim2022flame} can be characterized as caption-to-motion, where the models take in a pure text description of the desired motion.
While these methods can generate motions from textual descriptions, they are often limited in supporting instructions from users like InstructGPT~\cite{ouyang2022instructgpt}. In other words, they do not allow users to provide context-specific instructions for certain applications.
MotionCLIP~\cite{tevet2022motionclip} utilizes the language and visual understanding of CLIP~\cite{radford2021clip} to align its latent space with a motion auto-encoder.
Meanwhile, many language models, such as T5\cite{raffel2020t5} and InstructGPT~\cite{ouyang2022instructgpt}, have been developed to address diverse language processing tasks, including translation, question answering, and classification.
These models are typically designed to map a given text input to a target output, such as a translation or answer. 
However, while these models have shown remarkable performance in language tasks, they have not been widely applied to motion tasks. Therefore, we propose MotionGPT to enable the effective integration of natural language models with human motion tasks, providing a unified solution for motion synthesis problems.

\begin{figure}[t]
\centering
\includegraphics[width=\linewidth]{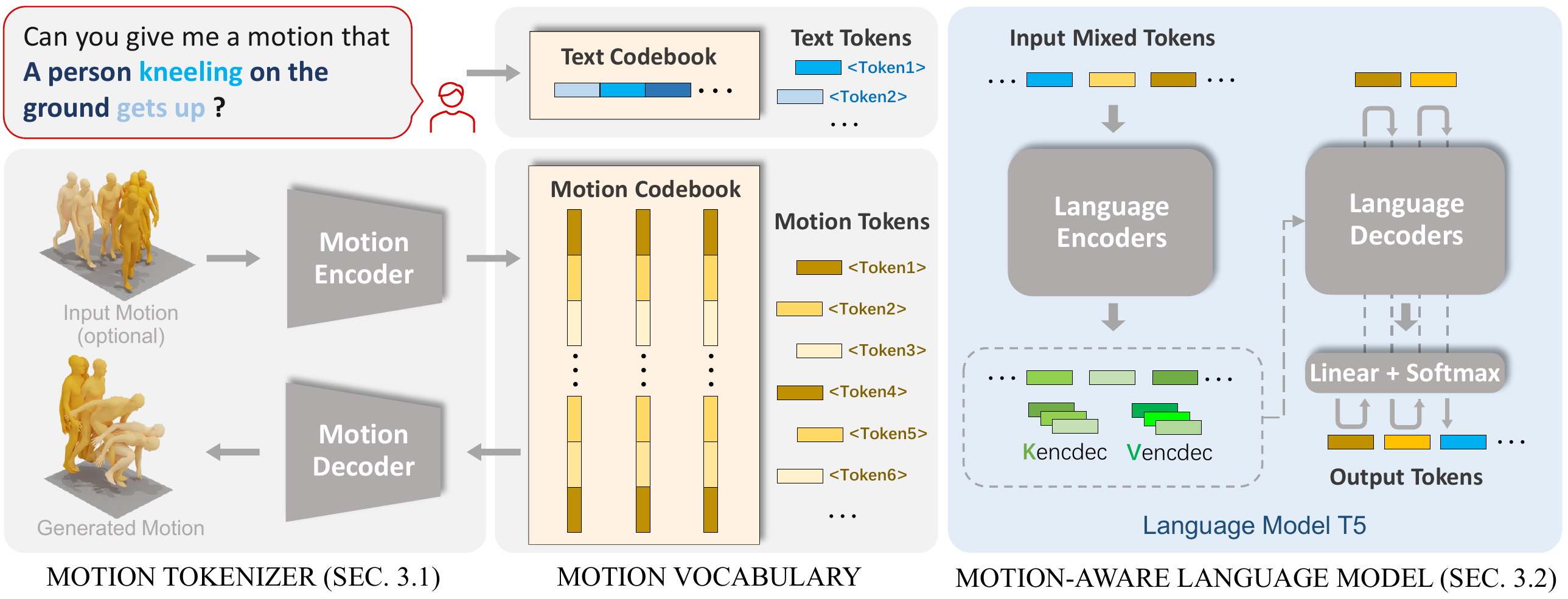}
\vspace{-10pt}
\caption{Method overview: MotionGPT consists of a motion tokenizer $\mathcal{V}$ (\cref{sec:method:vqvae}) and a motion-aware language model (\cref{sec:method:lm}). Combining \textit{Motion Tokens} learned by $\mathcal{V}$ and \textit{Text Tokens} by text tokenizer, we then learn motion and language jointly utilizing language model as backbone.}
\label{fig:pipeline}
\end{figure}

\section{Method}
\label{method}
To involve large language data and models in the motion generation tasks, we propose a unified motion-language framework named MotionGPT. As illustrated in \cref{fig:pipeline}, MotionGPT consists of a motion tokenizer responsible for converting raw motion data into discrete motion tokens (\cref{sec:method:vqvae}), as well as a motion-aware language model that learns to understand the motion tokens from large language pre-training \ct{models} by corresponding textual descriptions (\cref{sec:method:lm}). To address motion-relevant tasks, we introduce a three-stage training scheme (\cref{sec:method: strategy}) of MotionGPT for the training of motion tokenizer, motion-language pre-training, and instruction tuning.

We first propose the motion tokenizer consisting of a motion encoder $\mathcal{E}$ and a motion decoder $\mathcal{D}$, to encode a $M$ frame motion ${m}^{1:M}=\{{x}^i\}_{i=1}^{M}$ into $L$ motion tokens ${z}^{1:L}=\{{z}^i\}_{i=1}^{L}, L=M/l$, and decode ${z}^{1:L}$ back into the motion $\hat{m}^{1:M} = \mathcal{D}(z^{1:L}) = \mathcal{D}(\mathcal{E}(m^{1:M}))$, where $l$ denotes the temporal downsampling rate on motion length. Then, given an $N$ length sentence ${w}^{1:N}=\{w^i\}_{i=1}^{N}$ describing a motion-related question or demand, MotionGPT aims to generate its answer as $L$ length tokens $\hat{x}^{1:L}=\{\hat{x}^i\}_{i=1}^{L}$. It could be the human motion tokens $\hat{x}_m^{1:L}$ or the text tokens $\hat{x}_t^{1:L}$, which results in a motion ${\hat{m}}^{1:M}$ or a sentence $\hat{{w}}^{1:L}$
like a description of the given motion.

\subsection{Motion Tokenizer}
\label{sec:method:vqvae}

To represent motion in discrete tokens, we pre-train a 3D human motion tokenizer $\mathcal{V}$ based on the Vector Quantized Variational Autoencoders (VQ-VAE) architecture used in \cite{van2017vqvae,siyao2022bailando,chuan2022tm2t,zhang2023generating}. 
Our motion tokenizer consists of an encoder $\mathcal{E}$ and a decoder $\mathcal{D}$. The encoder generates discrete motion tokens with high informative density, while the decoder is able to reconstruct the motion tokens into motion sequences $\hat{m}^{1:M}$. This approach enables us to efficiently represent motion as a language, facilitating the integration of motion and language for various motion-related tasks.

Specifically, the motion encoder $\mathcal{E}$ first applies 1D convolutions to given frame-wise motion features $m^{1:M}$ along the time dimension, to obtain latent vectors $\hat{z}^{1:L}=\mathcal{E}(m^{1:M})$. 
Next, we transform $\hat{z}$ into a collection of codebook entries $z$ through discrete quantization. The learnable codebook $Z=\{{z}^i\}_{i=1}^{K} \subset \mathbb{R}^{d}$ consists of $K$ latent embedding vectors, each of dimension $d$. The process of quantization $Q(\cdot)$ replaces each row vector $b$ with its nearest codebook entry $b_k$ in $Z$, written as 
\begin{equation}
z_i=Q(\hat{z}^i):={\arg \min }_{z_k \in Z}\left\|\hat{z_i}-z_k\right\|_2.
\end{equation}
After quantization, the motion decoder $D$ project ${z}^{1:L}=\{{z}^i\}_{i=1}^{L}$ back to raw motion space as the motion $\hat{m}^{1:M}$ with $M$ frames. 
To train this motion tokenizer, we follow \cite{chuan2022tm2t,zhang2023generating} to utilize three distinct loss functions for training and optimizing the motion tokenizer: $ \mathcal{L}_\mathcal{V} = \mathcal{L}_{r} + \mathcal{L}_{e} + \mathcal{L}_{c}\label{eq:loss:vq},$ where the reconstruction loss $\mathcal{L}_{r}$, the embedding loss $\mathcal{L}_{e}$, and the commitment loss $\mathcal{L}_{c}$. To further improve the generated motion quality, we follow \cite{zhang2023generating} to utilize L1 smooth loss and velocity regularization in the reconstruction loss, as well as exponential moving average (EMA) and codebook reset techniques~\cite{razavi2019vqvae2} to enhance codebook utilization during training. We provide more details about the architecture and the training of our motion tokenizer in the supplement.
    
\subsection{Motion-aware Language Model}
\label{sec:method:lm}
Employing this motion tokenizer, a human motion $m^{1:M}$ can be mapped to a sequence of motion tokens $z^{1:L}$, allowing for joint representation with similar vocabulary embedding in language models~\cite{kudo2018sentencepiece,raffel2020t5,ouyang2022instructgpt}. By combining them in the unified vocabulary, we then learn motion and language jointly. 
%
We first represent motion tokens $z^{1:L}$ as a sequence of indices ${s}^{1:L}=\{{s}^i\}_{i=1}^{L}$, where ${s}^i$ corresponds to the index number of motion tokens ${z}^{1:L}$. On the other hand, previous language models, such as T5~\cite{raffel2020t5}, encode text as WordPiece tokens. They utilized a vocabulary of $K_t$ word pieces and trained the SentencePiece~\cite{kudo2018sentencepiece} model on a mixture of language datasets.

Most previous text-to-motion~\cite{chuan2022tm2t, chen2023mld, zhang2023generating} or motion-to-text~\cite{chuan2022tm2t} approaches employ different modules to handle text and motion individually, while we aim to model text and human motion \ct{together and in the same way.} 
To achieve this, we combine the original text vocabulary $V_t=\{{v}_t^i\}_{i=1}^{K_t}$ with motion vocabulary $V_m=\{{v}_m^i\}_{i=1}^{K_m}$, which is order-preserving to our motion codebook $Z$. 
Moreover, $V_m$ includes several special tokens like boundary indicators, for example, \textcolor{orange}{</som>} and \textcolor{orange}{</eom>} as the start and end of the motion.
Thus, we employ a new unified text-motion vocabulary $V = \{V_t, V_m\}$, and can formulate diverse motion-related tasks in a general format, where both input "words" and output "words" are from the same $V$.
These "words" can represent natural language, human motion, or even a mixture of two, depending on the specific task to be solved.
Therefore, our MotionGPT allows for the flexible representation and generation of diverse motion-related outputs within a single model.


To address the conditioned generation task, we employ a transformer-based model based on the architecture proposed in \cite{raffel2020t5}, which effectively maps the input sequences to the output.
Our source input consists of a sequence of tokens ${X_s}=\{{x_s}^i\}_{i=1}^{N}$, where $x_s\in V$ and $N$ represents the input length. Similarly, the target output is ${X_t}=\{{x_t}^i\}_{i=1}^{L}$, where $x_t\in V$ and $L$ denotes the output length.
As shown in \cref{fig:pipeline}, the source tokens are fed into the transformer encoder, and the subsequent decoder predicts the probability distribution of the potential next token at each step $p_\theta(x_t \mid x_s)=\prod_i p_\theta\left(x_t^i \mid x_t^{<i}, x_s\right)$ in an autoregressive manner. Therefore, during the training process, the objective is to maximize the log-likelihood of the data distribution:
\begin{equation}
    \mathcal{L}_{LM}=-\sum_{i=0}^{L_t-1} \log p_\theta\left(x_t^i \mid x_t^{<i}, x_s\right) .
    \label{eq:loss:lm}
\end{equation}
By optimizing this objective, MotionGPT learns to capture the underlying patterns and relationships from the data distribution, facilitating the accurate and meaningful generation of the target "words".
During the inference process, the target tokens are sampled recursively from the predicted distribution $p_\theta\left(\hat{x_t}^i \mid \hat{x_t}^{<i}, x_s\right)$ until the end token (i.e., \textcolor{NavyBlue}{</s>}). This sampling strategy enables the generation of the target sequence in a step-by-step manner, where each token is probabilistically determined based on the previously generated tokens and the given source input.

\subsection{Training Strategy}
\label{sec:method: strategy}
\begin{figure*}[t]
	\centering
	\includegraphics[width=\linewidth]{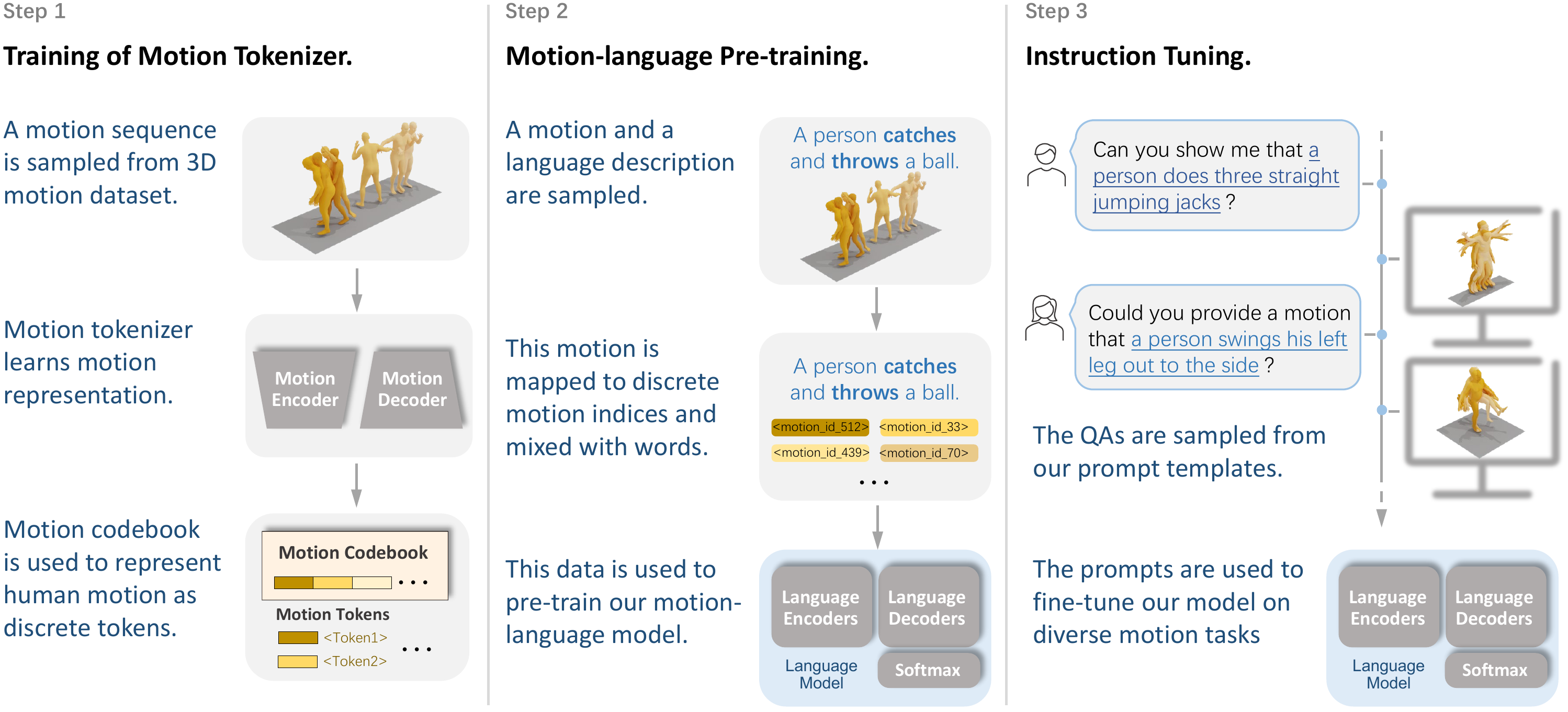}
\vspace{-18pt}
\caption{ Training Scheme. We introduce three training steps for our MotionGPT (\cref{sec:method: strategy}): First $\mathcal{V}$ learn a codebook for discrete motion representation. Then we train language using a mixture of language and motion data to learn the semantic coupling between text and motion. Finally, we fine-tune the model in a multi-task text-motion dataset with instructions.}
	\label{fig:training}
 \vspace{-15pt}
\end{figure*}

Since T5s have only been exposed to language data, represented within a text vocabulary $V_t$, we thus bridge motion and language and enable this language model to comprehend human motion concepts, by learning the motion vocabulary $V_m$. As shown in \cref{fig:training}, our training scheme includes three stages: (1) Training of motion tokenizer, which focuses on learning the motion codebook to represent human motion as discrete tokens. (2) Motion-language pre-training stage, which includes unsupervised and supervised objectives to learn the relationship between motion and language. (3) Instruction tuning stage, \ct{which tunes the model based on} prompt-based instructions for different motion-relevant tasks.

\textbf{Training of Motion Tokenizer.} We first learn the motion tokenizer using the objective defined in Equation \ref{eq:loss:vq}. This training process allows any human motion sequence $\hat{x}^{1:L}$ to be represented as a sequence of motion tokens, enabling seamless integration with textual information. Once optimized, the motion tokenizer remains unchanged throughout the subsequent stages of the pipeline.

\textbf{Motion-language Pre-training Stage.} The T5 models \cite{raffel2020t5,chung2022flant5} are trained and fine-tuned on natural language datasets with instruction-based phrasing \cite{chung2022flant5,ouyang2022instructgpt}. We continue to pre-train this model using a mixture of language and motions data in both unsupervised and supervised manners:
1) To generalize to various downstream tasks like~\cite{devlin2018bert,radford2019gpt2,raffel2020t5,ouyang2022instructgpt}, we follow \cite{raffel2020t5} to design an objective, where a certain percentage (15\%) of tokens in the input tokens $X_s$ are randomly replaced with a special sentinel token.
%
%
On the other side, the corresponding target sequence is constructed by extracting the dropped-out spans of tokens, delimited by the same sentinel tokens used in the input sequence, along with an additional sentinel token to indicate the end of the target sequence.
2) We then learn the motion-language relation by the supervision of paired text-motion datasets \cite{Guo_2022_CVPR_humanml3d,Plappert2016kit}. We train MotionGPT on the supervised motion-language translation, where the input is either a human motion or a text description.
After unsupervised and supervised training processes, we aim to equip our model with the understanding of text and motion relationships.

\textbf{Instruction Tuning Stage.} We construct a multi-task text-motion dataset by formulating it as instructions, building upon the foundation of existing text-to-motion datasets such as HumanML3D \cite{Guo_2022_CVPR_humanml3d} and KIT \cite{Plappert2016kit}. Specifically, we define 15 core motion tasks, such as motion generation with text, motion captioning, motion prediction, and others. For each task, we compose dozens of different instruction templates, resulting in more than one thousand different tasks, each having a unique instruction prompt.
For example, an instruction prompt for motion generation task could be ``\textcolor{NavyBlue}{
Can you generate a motion sequence that depicts `a person emulates the motions of a waltz dance'?}''. Similarly, for the motion captioning task, the instruction prompt could be ``\textcolor{NavyBlue}{Provide an accurate caption describing the motion of \textcolor{orange}{<motion\_tokens>}}'', where \textcolor{orange}{<motion\_tokens>} represents a sequence of motion tokens generated by our motion tokenizer.
We have demonstrated the efficacy of instruction tuning in \cref{sec:ablation:strategy}, which leads to improvement across various tasks and enhances the model performance for unseen tasks or prompts. More examples of prompts are provided in the supplements.

\section{Experiments}
\vspace{-10pt}
\label{experiments}
Extensive comparisons evaluate the performance of our MotionGPTs across multiple motion-relevant tasks and datasets. 
Details of the dataset settings, evaluation metrics, and implementation specifics (\cref{sec:comp:detail}) are provided.
We first present a uniform benchmark by comparing our approach with other SOTAs across various tasks (\cref{sec:comp:general}). Then, we evaluate each specific comparison on text-to-motion (\cref{sec:comp:text}), motion-to-text (\cref{sec:comp:m2t}), motion prediction and motion in-between (\cref{sec:comp:motion}).
The supplements include more qualitative results, user studies, and further implementation details.

\begin{table}[t]
\resizebox{\columnwidth}{!}{%
\begin{tabular}{@{}lcccccccccc@{}}
\toprule
\multirow{2}{*}{Methods} & 
\multicolumn{3}{c}{Text-to-Motion}& 
\multicolumn{3}{c}{Motion-to-Text}& 
\multicolumn{2}{c}{Motion Prediction}&
\multicolumn{2}{c}{Motion In-between}
\\ \cmidrule(lr){2-4} \cmidrule(lr){5-7} \cmidrule(lr){8-9} \cmidrule(lr){10-11}
&
R TOP1$\uparrow$ 
 & FID$\downarrow$ 
&  
DIV$\rightarrow$ 
& R TOP3$\uparrow$ 
& Bleu@4$\uparrow$ & Cider$\uparrow$  & FID$\downarrow$ & DIV$\rightarrow$  & FID$\downarrow$ & DIV$\rightarrow$
\\ \midrule
Real &
 $0.511^{\pm.003}$ &
  $0.002^{\pm.000}$ &
  $9.503^{\pm.065}$ &
  
  $0.828$ 
  & \multicolumn{1}{c}{-}& \multicolumn{1}{c}{-}
  & 0.002 & 9.503 & 0.002 & 9.503
  \\ \midrule
MLD \cite{chen2023mld} &
 $0.481^{\pm.003}$ &
  ${0.473}^{\pm.013}$ &

  $9.724^{\pm.082}$ &
  
  \multicolumn{1}{c}{-}& \multicolumn{1}{c}{-}& \multicolumn{1}{c}{-}& \multicolumn{1}{c}{-}& \multicolumn{1}{c}{-}& \multicolumn{1}{c}{-} &
  \multicolumn{1}{c}{-}
  \\
T2M-GPT \cite{mdm2022human}&
   $\underline{0.491}^{\pm.003}$&
  $\boldsymbol{0.116}^{\pm.004}$ &
  $9.761^{\pm.081}$ &
  \multicolumn{1}{c}{-}& \multicolumn{1}{c}{-}& \multicolumn{1}{c}{-} & \multicolumn{1}{c}{-}& \multicolumn{1}{c}{-} & \multicolumn{1}{c}{-}& \multicolumn{1}{c}{-}
  \\
TM2T \cite{chuan2022tm2t} &
 $0.424^{\pm.017}$ &
  ${1.501^{\pm.003}}$ &
  $8.589^{\pm.076}$ &
  $0.823$&
  ${7.00}$ & 
  ${16.8}$& \multicolumn{1}{c}{-}& \multicolumn{1}{c}{-}& \multicolumn{1}{c}{-}& \multicolumn{1}{c}{-}
  \\
MDM \cite{mdm2022human}&
 $0.320^{\pm005}$ &
${0.544}^{\pm.044}$ &
  $\underline{9.559}^{\pm.086}$ &
  \multicolumn{1}{c}{-}& \multicolumn{1}{c}{-}& \multicolumn{1}{c}{-}& $6.031$ & $7.813$& $2.698 $&$ 8.420$
  \\
 \midrule
MotionGPT (Ours) &
$\boldsymbol{0.492}^{\pm.003}$ &
$\underline{0.232}^{\pm.008}$ & 
$\boldsymbol{9.528}^{\pm.071}$ &
$\boldsymbol{0.827}$
& $\boldsymbol{12.47}$ & $\boldsymbol{29.2}$ & $\boldsymbol{0.905}$ &$\boldsymbol{8.972}$  & $\boldsymbol{0.214}$&$\boldsymbol{9.560}$
   \\ \bottomrule
\end{tabular}%
}
\caption{Comparison of four motion-related tasks on HumanML3D~\cite{Guo_2022_CVPR_humanml3d} dataset. The evaluation metrics are computed using the encoder introduced in \cite{Guo_2022_CVPR_humanml3d}. The empty columns of previous methods indicate that they can not handle the task. The arrows ($\rightarrow$) indicate that closer to \textit{Real} is desirable. \textbf{Bold} and \underline{underline} indicate the best and the second best result on text-to-motion task.}
\label{tab:uniform}
\end{table}

\subsection{Experimental Setup}
\label{sec:comp:metric}
\textbf{Datasets.} General motion synthesis can support diverse task settings, and thus previous datasets and a modified benchmark are utilized to evaluate MotionGPT.
The study primarily focuses on two text-to-motion datasets: HumanML3D~\cite{Guo_2022_CVPR_humanml3d} and KIT \cite{Plappert2016kit}. 
The KIT dataset provides 6,353 textual descriptions corresponding to 3,911 motion sequences, while the HumanML3D dataset \cite{Guo_2022_CVPR_humanml3d} is a more recent dataset that contains 14,616 motion sequences obtained from AMASS \cite{AMASS_ICCV2019}, along with 44,970 sequence-level textual descriptions. 
To evaluate MotionGPT as a uniform framework on tasks, such as motion prediction and motion completion (in-between), we utilize the motion sequences available in HumanML3D, which is also a subset of the larger AMASS dataset.
Following the previous works \cite{Guo_2022_CVPR_humanml3d, chen2023mld, mdm2022human}, we adopt the same motion representation for fair comparisons, which combines joint velocities, positions, and rotations. 
By using this consistent representation, MotionGPT enables the availability to support further studies in the field.
($cf.$ supplement for the benchmark details.)

\textbf{Evaluation Metrics} are summarized as four parts. (1) Motion quality: 
Frechet Inception Distance (FID) is our primary metric based on a feature extractor \cite{Guo_2022_CVPR_humanml3d} to evaluate the distance of feature distributions between the generated and real motions.
For motion completion, we utilize metrics used in motion prediction studies \cite{yuan2020dlow, zhang2021we, ma2022multi}, such as Average Displacement Error (ADE) and Final Displacement Error (FDE), to evaluate the accuracy of the predicted motion.
%
%
(2) Generation diversity: 
We utilize the Diversity (DIV) metric to assess the motions diversity, which calculates the variance through features extracted from the motions \cite{Guo_2022_CVPR_humanml3d}. MultiModality (MM) measures the diversity of generated motions within the same text description of motion.
(3) Text matching: Based on the feature space from \cite{Guo_2022_CVPR_humanml3d}, the motion-retrieval precision (R Precision) evaluates the accuracy of matching between texts and motions using Top 1/2/3 retrieval accuracy. Multi-modal Distance (MM Dist) measures the distance between motions and texts.
(4) Linguistic quality: We follow \cite{chuan2022tm2t} utilizing linguistic metrics from natural language studies, including BLUE~\cite{papineni2002bleu}, Rouge~\cite{lin2004rouge}, Cider~\cite{vedantam2015cider}, and BertScore~\cite{zhang2019bertscore} to evaluate the quality of generated motion captions.

\label{sec:comp:detail}
\textbf{Implementation Details.} We set the codebook of motion tokenizer as $K\in\mathbb{R}^{512\times512}$ for most comparisons. The motion encoder $\mathcal{E}$ incorporates a temporal downsampling rate $l$ of 4. 
We utilize T5 \cite{raffel2020t5} as the underlying architecture for our language model, with a baseline model consisting of 12 layers in both the transformer encoder and decoder. The feed-forward networks have an output dimensionality of $d_\text{ff} = 3072$, and the attention mechanisms employ an inner dimensionality of $d_\text{kv} = 64$. The remaining sub-layers and embeddings have a dimensionality of $d_\text{model} = 768$.
%
%
Moreover, all our models employ the AdamW optimizer for training. The motion tokenizers are trained utilizing a $10^{-4}$ learning rate and a 256 mini-batch size, while our language models have a $2\times10^{-4}$ learning rate for the pre-train stage, $10^{-4}$ for the instruction tuning stage, and a 16 mini-batch size for both stages. The motion tokenizer undergoes 150K iterations of training, while the language model undergoes 300K iterations during the pre-train stage and another 300K iterations during the instruction tuning stage. All models are trained on 8 Tesla V100 GPUs.


\subsection{Comparisons on Motion-relevant Tasks}
\label{sec:comp}
\label{sec:comp:general}
\textbf{Comparisons on Multiple Tasks.} By introducing a uniform framework that treats human motion as a foreign language, we open up the exploration of diverse motion-relevant tasks. We employ a 220M pre-trained \textit{Flan-T5-Base}\cite{raffel2020t5,chung2022flant5} model as our backbone and fine-tune the model through the pre-training and instruction tuning stage (\cref{sec:method: strategy}) for all following comparisons.
As shown in \cref{tab:uniform}, we evaluate MotionGPT against state-of-the-art methods on key tasks such as text-conditioned motion generation~\cite{chen2023mld,zhang2023generating,chuan2022tm2t,mdm2022human}, motion captioning~\cite{chuan2022tm2t}, motion prediction~\cite{mdm2022human}, and motion in-between\cite{mdm2022human}. 
While we leverage existing results from previous works or benchmarks for text-to-motion and motion-to-text tasks, we re-implement the motion diffusion models \cite{mdm2022human} for motion prediction and evaluate it under the same metrics and settings.
Please note that some methods are designed for specific tasks, and thus some metrics are empty for tasks they cannot handle.
The results presented in \cref{tab:uniform} demonstrate that our MotionGPT achieves competitive performance across all evaluated tasks, highlighting its capability to address diverse motion tasks within a single model.

\label{sec:comp:text}
\textbf{Comparisons on Text-to-Motion.} The text-to-motion task involves generating human motion sequences based on a given text input. We evaluate the proposed the MotionGPT model as the pre-trained MotionGPT, the same one in \cref{tab:uniform}, as well as fine-tuned it on text-to-motion task. We compare our MotionGPTs with other SOTAs~\cite{chuan2022tm2t,Guo_2022_CVPR_humanml3d,mdm2022human,chen2023mld,zhang2023generating} and evaluate the performance on both HumanML3D and KIT datasets using suggested metrics \cite{Guo_2022_CVPR_humanml3d}. The results are computed with a 95\% confidence interval, obtained from 20 repeated runs.
The majority of the reported results are taken directly from their own papers or the benchmark presented in \cite{Guo_2022_CVPR_humanml3d}.
\cref{tab:tm:comp:humanml3d} summarizes the comparison results, where MotionGPT achieves competitive performance on most metrics.

\begin{table}[t]
\centering
\vspace{10pt}
\resizebox{\columnwidth}{!}{%
\begin{tabular}{@{}lccccccc@{}}
\toprule
\multirow{2}{*}{Methods}&\multicolumn{3}{c}{RPrecision$\uparrow$}&\multicolumn{1}{c}{\multirow{2}{*}{FID$\downarrow$}}&\multirow{2}{*}{MMDist$\downarrow$}&\multirow{2}{*}{Diversity$\rightarrow$}&\multirow{2}{*}{MModality$\uparrow$}\\\cmidrule(lr){2-4}
&\multicolumn{1}{c}{Top1}&\multicolumn{1}{c}{Top2}&\multicolumn{1}{c}{Top3}&\multicolumn{1}{c}{}&&&\\\midrule
Real &
  $0.511^{\pm.003}$ &
  $0.703^{\pm.003}$ &
  $0.797^{\pm.002}$ &
  $0.002^{\pm.000}$ &
  $2.974^{\pm.008}$ &
  $9.503^{\pm.065}$ &
  \multicolumn{1}{c}{-}
  \\ \midrule
TM2T \cite{chuan2022tm2t} &
  $0.424^{\pm.003}$ &
  $0.618^{\pm.003}$ &
  $0.729^{\pm.002}$ &
  $1.501^{\pm.017}$ &
  $3.467^{\pm.011}$ &
  $8.589^{\pm.076}$ &
  $2.424^{\pm.093}$ 
  \\
T2M \cite{Guo_2022_CVPR_humanml3d}&
  $0.457^{\pm.002}$ &
  $0.639^{\pm.003}$ &
  $0.740^{\pm.003}$ &
  $1.067^{\pm.002}$ &
  $3.340^{\pm.008}$ &
  $9.188^{\pm.002}$ &
  $2.090^{\pm.083}$ \\
MotionDiffuse \cite{zhang2022motiondiffuse} &
$\underline{0.491}^{\pm.001}$ &
$\boldsymbol{0.681}^{\pm.001}$ &
$\boldsymbol{0.782}^{\pm.001}$ &
$0.630^{\pm.001}$ &
$\underline{3.113}^{\pm.001}$ &
${9.410}^{\pm.049}$ &
$1.553^{\pm.042}$ \\
MDM \cite{mdm2022human}&
  $0.320^{\pm.005}$ &
  $0.498^{\pm.004}$ &
  $0.611^{\pm.007}$ &
  ${0.544}^{\pm.044}$ &
  $5.566^{\pm.027}$ &
  $\underline{9.559}^{\pm.086}$ &
  $\underline{2.799}^{\pm.072}$ \\
MLD \cite{chen2023mld}&
  ${0.481}^{\pm.003}$ &
  ${0.673}^{\pm.003}$ &
  ${0.772}^{\pm.002}$ &
  ${0.473}^{\pm.013}$ &
  ${3.196}^{\pm.010}$ &
  $9.724^{\pm.082}$ &
  ${2.413}^{\pm.079}$
  \\
T2M-GPT \cite{zhang2023generating}  & $\underline{0.491}^{\pm.003}$ & ${0.680}^{\pm.003}$ & ${0.775}^{\pm.002}$ & $\boldsymbol{0.116}^{\pm.004}$ & ${3.118}^{\pm.011}$ & $9.761^{\pm.081}$ & $1.856^{\pm.011}$ \\
\midrule
MotionGPT (Pre-trained) &
${0.435}^{\pm.003}$ & $0.607^{\pm.002}$ & $0.700^{\pm.002}$ & $\underline{0.160}^{\pm.008}$ & $3.700^{\pm.009}$ & $9.411^{\pm.081}$ & $\boldsymbol{3.437}^{\pm.091}$
\\
MotionGPT (Fine-tuned) &
$\boldsymbol{0.492}^{\pm.003}$ & $\boldsymbol{0.681}^{\pm.003}$ & $\underline{0.778}^{\pm.002}$ & $0.232^{\pm.008}$ & $\boldsymbol{3.096}^{\pm.008}$ & $\boldsymbol{9.528}^{\pm.071}$ & 
${2.008}^{\pm.084}$

   \\ \bottomrule
\end{tabular}%
}
\vspace{5pt}
\caption{Comparison of text-to-motion on HumanML3D~\cite{Guo_2022_CVPR_humanml3d}. The empty MModality indicates \textit{Real} motion is deterministic. These methods are sorted by FID. \textit{Pre-trained} and \textit{Fine-tuned} indicate uniform motion-language pre-training and specific fine-tuning on this task. ($cf.$ \cref{tab:uniform} for notations.)}
\vspace{-5pt}
\label{tab:tm:comp:humanml3d}
\end{table}

\label{sec:comp:m2t}
\textbf{Comparisons on Motion-to-Text.}
The motion-to-text task involves generating a text description based on a given human motion sequence. 
We compare the pre-trained MotionGPT with recent work TM2T~\cite{chuan2022tm2t}. We evaluate the performance on the HumanML3D using the suggested metrics from \cite{chuan2022tm2t}. Additionally, we measure the average numbers of words $\text{Length}_\text{avg}$ for further comparisons. Please note that the reported results in \cite{chuan2022tm2t} are evaluated with pre-processed ground truth text, which ignores the grammatical tense and plural forms of words. In \cref{tab:tm:comp:m2t}, we directly use the ground truth text descriptions for a more accurate assessment. This comparison shows that MotionGPT overperforms recent work on text descriptions of given motions.

\begin{table}[t]
\resizebox{\columnwidth}{!}{%
\begin{tabular}{@{}lccccccccc@{}}
\toprule
\multirow{2}{*}{Methods}&\multicolumn{2}{c}{RPrecision$\uparrow$}&
\multirow{2}{*}{MMDist$\downarrow$} &\multirow{2}{*}{$\text{Length}_{\text{avg}}$$\uparrow$} &\multirow{2}{*}
{Bleu@1$\uparrow$}&
\multirow{2}{*}{Bleu@4$\uparrow$}&\multirow{2}{*}{Rouge$\uparrow$}&\multirow{2}{*}{Cider$\uparrow$}&\multirow{2}{*}{BertScore$\uparrow$}
\\\cmidrule(lr){2-3}
&\multicolumn{1}{c}{Top1}&\multicolumn{1}{c}{Top3}&&&\\\midrule

Real & $0.523$ 
& $0.828$ & $2.901$ &  $12.75$ &\multicolumn{1}{c}{-}& \multicolumn{1}{c}{-}& \multicolumn{1}{c}{-}& \multicolumn{1}{c}{-}& \multicolumn{1}{c}{-}
\\ \midrule
TM2T\cite{chuan2022tm2t} & 
${0.516}$ & 
${0.823}$  &${2.935}$ & $10.67$& $\boldsymbol{48.9}$ & $7.00$ & $\boldsymbol{38.1}$ & $16.8$ & ${32.2}$ \\
MotionGPT (Ours)& 
$\boldsymbol{0.543}$ & $\boldsymbol{0.827}$ & $\boldsymbol{2.821}$ & $\boldsymbol{13.04}$ 
& $48.2$ & $\boldsymbol{12.47}$ & $37.4$ & $\boldsymbol{29.2}$ & $\boldsymbol{32.4}$
\\
   \bottomrule
\end{tabular}%
}
\vspace{5pt}
\caption{Comparison of motion captioning on HumanML3D~\cite{Guo_2022_CVPR_humanml3d}. The evaluation metrics follow \cite{chuan2022tm2t}, while we use the ground truth texts without pre-processing for linguistic metrics calculation.}
\vspace{-15pt}
\label{tab:tm:comp:m2t}
\end{table}

\textbf{Comparisons on Motion Prediction and In-between.}
\label{sec:comp:motion}
We summarize motion prediction and in-between together as general motion completion.
To evaluate the motion completion capability of MotionGPT, we employ part of the AMASS dataset \cite{AMASS_ICCV2019}, a motion-only dataset. 
For motion prediction task, we only input around the first 20\% of the motion sequence as conditions. For in-between, we mask about 50\% motion randomly for completion.
We also fine-tune MotionGPT specifically for this task and employ FID, ADE, and FDE as metrics like \cref{sec:comp:metric}.
Furthermore, we evaluate MDM~\cite{mdm2022human} on motion prediction by utilizing their provided model, which also supports motion in-between through masked motion ``in-painting''.
The real motion data is used as one of our baselines.
\cref{tab:comp:motion} reports that our MotionGPT has the best motion completion quality and diversity.

\begin{table}[h]
\centering
\resizebox{0.8\columnwidth}{!}{%
\begin{tabular}{@{}lccccccccc@{}}
\toprule
\multirow{2}{*}{Methods} & 
\multicolumn{4}{c}{Motion Prediction }& &
\multicolumn{3}{c}{Motion In-between  }\\
\cmidrule(lr){2-5} \cmidrule(lr){7-9}
& $\text{FID}\downarrow $  
& Diversity$\uparrow$            
& ADE$\downarrow$
& FDE$\downarrow$
&& $\text{FID}\downarrow $  
& Diversity$\uparrow$            
& ADE$\downarrow$
\\ 
\toprule
Real & $0.002$ & $9.503$ & - & - && $0.002$ & $9.503$ & -
\\\midrule
MDM\cite{mdm2022human}& $6.031$ & $7.813$& $5.446$ &$8.561$&& $2.698 $&$ 8.420$& $3.787$
\\
MotionGPT (Ours) 
& $\boldsymbol{0.905}$ &$\boldsymbol{8.972}$ & $\boldsymbol{4.745}$& $\boldsymbol{6.040}$ &&$\boldsymbol{0.214}$&$\boldsymbol{9.560}$&$\boldsymbol{3.762}$
\\ \bottomrule
\end{tabular}%
}
\vspace{4pt}
\caption{Comparison of motion prediction and motion in-between on part of AMASSS~\cite{AMASS_ICCV2019} dataset using motion data only. 
FID indicates motion quality and Diversity (DIV) for motion diversity within each condition. ADE and FDE are joints distance between generation and ground truth.}
\vspace{-10pt}
\label{tab:comp:motion}
\end{table}

\subsection{Ablation Studies}
\label{sec:ablation}
MotionGPT employs T5~\cite{raffel2020t5} as the motion-aware language backbone model, and we train these models with pre-training and then instruction tuning. Thus, both model size and training strategy influence the performance of MotionGPTs. We here evaluate them on the typical motion tasks. More detailed ablation studies are provided in the supplements.

\begin{table}[t]
\resizebox{\columnwidth}{!}{%
\begin{tabular}{@{}lccccccccccc@{}}
\toprule
\multirow{2}{*}{Size} & 
\multirow{2}{*}{Instruction Tuning} & 
\multicolumn{3}{c}{Text-to-Motion}& 
\multicolumn{3}{c}{Motion-to-Text} & 
\multicolumn{2}{c}{Motion Prediction}&
\multicolumn{2}{c}{Motion In-between}
\\ \cmidrule(lr){3-5} \cmidrule(lr){6-8} 
\cmidrule(lr){9-10} \cmidrule(lr){11-12}
&& R TOP3 $\uparrow$ & FID $\downarrow$ & DIV $\rightarrow$ & MMDist$\downarrow$ & Bleu@4$\uparrow$ & Cider$\uparrow$ 
&   FID $\downarrow$ & DIV $\rightarrow$  & FID $\downarrow$ & DIV $\rightarrow$ 
\\ \midrule
Real & - &
$0.797\hidden{^{\pm.002}}$ &
$0.002\hidden{^{\pm.000}}$ &
$9.503\hidden{^{\pm.065}}$ &
${2.901}$
& -& -
& 0.002 & 9.503 & 0.002 & 9.503
\\ \midrule
Small &  &
$0.706\hidden{^{\pm.002}}$	& $0.727\hidden{^{\pm.008}}$	& 
$9.264\hidden{^{\pm.042}}$  &
$\boldsymbol{2.748}$	& ${12.02}$& ${24.9}$ & - & - & - & - 
\\
Small & \checkmark &
$0.663$	& ${0.336}$	& ${9.239}$ &
$2.931$	& $10.54$	& $24.3$ & $0.954$& $8.727$ & $0.326$ & $9.618$
\\ \midrule
Base & &
$\boldsymbol{0.722}\hidden{^{\pm.002}}$	& $0.365\hidden{^{\pm.008}}$	& 
$9.407\hidden{^{\pm.042}}$ &
$2.821$	& $\boldsymbol{12.47}$ & $\boldsymbol{29.2}$ & - & - & - & - 
\\
Base & \checkmark &
$0.700$	& $0.160$	& $\boldsymbol{9.411}$ &
$3.019$	& $11.42$ & $28.2$ 
& ${0.905}$ &${8.972}$&$\boldsymbol{0.214}$&$\boldsymbol{9.560}$
\\ \midrule
Large &  &
$0.694\hidden{^{\pm.002}}$ & $0.234\hidden{^{\pm.009}}$ & $9.310\hidden{^{\pm.078}}$ & $2.776$& $12.44$ & $28.5$ & - & - & - & - 
\\
Large & \checkmark &
$0.708$	& $\boldsymbol{0.159}$	& $9.301$ &
$3.011$ & $11.71$ & $29.1$ &
$\boldsymbol{0.556}$ & $\boldsymbol{8.975}$ & $0.223$  & $9.358$
\\ 
\bottomrule
\end{tabular}%
}
\vspace{2pt}
\caption{Evaluation of instruction tuning and different model sizes of MotionGPTs in four motion tasks on HumanML3D~\cite{Guo_2022_CVPR_humanml3d} dataset. ($cf.$ \cref{tab:uniform} for metrics details)}
\vspace{-15pt}
\label{tab:abl:tuning}
\end{table}

\label{sec:ablation:motionsize}
\textbf{Model Sizes.}
We evaluate the performance of models with different sizes across four motion tasks.
Besides the base 220M MotionGPT in \cref{sec:comp:detail}, we now evaluate 60M, 220M, and 770M MotionGPTs. \cref{tab:abl:tuning} demonstrates that the 220M base model has achieved remarkable performance compared to the smaller 60M model. However, the larger model size of current Motions does not yield significant improvements and, in few cases, even leads to worse results, as observed in the motion in-between task. We believe this could be caused by the small amount of current motion datasets. HumanML3D only includes 15k motion sequences, much smaller than even billions of language and image data.

\label{sec:ablation:strategy}
\textbf{Effectiveness of Instruction Tuning.}
We evaluate the impact of our instruction tuning strategy on different model sizes. The results in \cref{tab:abl:tuning} demonstrate that instruction tuning enhances the versatility of MotionGPT, enabling more motion tasks like motion completion and improving the motion performance of the text-to-motion task. However, for pure text-generation tasks, the model performance is downgraded, likely due to the pair amount of textual descriptions and coupled motions.

\section{Disscusion}
As the first trial, to our best knowledge, exploring human motion generation using language models, the proposed MotionGPT still owns limitations as follows.
MotionGPT only utilizes motion on articulated human bodies, while many other works focus on faces~\cite{karras2017audio, cao2018sparse}, hands~\cite{romero2022embodied, li2022nimble, li2021piano} and even animal~\cite{Rueeg:CVPR:2022, Zuffi:CVPR:2018} motion.
Besides, our method is also restricted to multiple humans without modeling human-object, or human-environment interactions~\cite{shafir2023priormdm}.
It is interesting to model the human interaction scenarios in a motion-language framework and generate controllable motions~\cite{shafir2023priormdm}.

We summarize the proposed MotionGPT as a uniform motion-language framework to generate plausible human motion and natural language descriptions through prompt-based instructions. Compared to the compatible motion diffusion methods~\cite{chen2023mld, mdm2022human}, our MotionGPT produces competitive results on motion generation, motion captioning, motion prediction, and motion in-between using only one pre-trained generative model. With the advancement of large language data and models~\cite{raffel2020t5,chung2022flant5}, MotionGPT is also capable of addressing natural question-to-answer tasks.
Extensive experiments on various human motion-relevant tasks demonstrate the effectiveness and extendibility of MotionGPT.



\newpage
{
\small
\bibliographystyle{plain}

\begin{thebibliography}{10}

  \bibitem{ahuja2019language2pose}
  Chaitanya Ahuja and Louis-Philippe Morency.
  \newblock Language2pose: Natural language grounded pose forecasting.
  \newblock In {\em 2019 International Conference on 3D Vision (3DV)}, pages
    719--728. IEEE, 2019.
  
  \bibitem{bach2022promptsource}
  Stephen~H Bach, Victor Sanh, Zheng-Xin Yong, Albert Webson, Colin Raffel,
    Nihal~V Nayak, Abheesht Sharma, Taewoon Kim, M~Saiful Bari, Thibault Fevry,
    et~al.
  \newblock Promptsource: An integrated development environment and repository
    for natural language prompts.
  \newblock {\em arXiv preprint arXiv:2202.01279}, 2022.
  
  \bibitem{brown2020gpt3}
  Tom Brown, Benjamin Mann, Nick Ryder, Melanie Subbiah, Jared~D Kaplan, Prafulla
    Dhariwal, Arvind Neelakantan, Pranav Shyam, Girish Sastry, Amanda Askell,
    et~al.
  \newblock Language models are few-shot learners.
  \newblock {\em Advances in neural information processing systems},
    33:1877--1901, 2020.
  
  \bibitem{cao2018sparse}
  Xuan Cao, Zhang Chen, Anpei Chen, Xin Chen, Shiying Li, and Jingyi Yu.
  \newblock Sparse photometric 3d face reconstruction guided by morphable models.
  \newblock In {\em Proceedings of the IEEE Conference on Computer Vision and
    Pattern Recognition}, pages 4635--4644, 2018.
  
  \bibitem{chung2022flant5}
  Hyung~Won Chung, Le~Hou, Shayne Longpre, Barret Zoph, Yi~Tay, William Fedus,
    Eric Li, Xuezhi Wang, Mostafa Dehghani, Siddhartha Brahma, et~al.
  \newblock Scaling instruction-finetuned language models.
  \newblock {\em arXiv preprint arXiv:2210.11416}, 2022.
  
  \bibitem{dai2019transformer}
  Zihang Dai, Zhilin Yang, Yiming Yang, Jaime Carbonell, Quoc~V Le, and Ruslan
    Salakhutdinov.
  \newblock Transformer-xl: Attentive language models beyond a fixed-length
    context.
  \newblock {\em arXiv preprint arXiv:1901.02860}, 2019.
  
  \bibitem{devlin2018bert}
  Jacob Devlin, Ming-Wei Chang, Kenton Lee, and Kristina Toutanova.
  \newblock Bert: Pre-training of deep bidirectional transformers for language
    understanding.
  \newblock {\em arXiv preprint arXiv:1810.04805}, 2018.
  
  \bibitem{girdhar2023imagebind}
  Rohit Girdhar, Alaaeldin El-Nouby, Zhuang Liu, Mannat Singh, Kalyan~Vasudev
    Alwala, Armand Joulin, and Ishan Misra.
  \newblock Imagebind: One embedding space to bind them all.
  \newblock {\em arXiv preprint arXiv:2305.05665}, 2023.
  
  \bibitem{goutsu2021seqgan}
  Yusuke Goutsu and Tetsunari Inamura.
  \newblock Linguistic descriptions of human motion with generative adversarial
    seq2seq learning.
  \newblock In {\em 2021 IEEE International Conference on Robotics and Automation
    (ICRA)}, pages 4281--4287. IEEE, 2021.
  
  \bibitem{gower1975generalized}
  John~C Gower.
  \newblock Generalized procrustes analysis.
  \newblock {\em Psychometrika}, 40(1):33--51, 1975.
  
  \bibitem{Guo_2022_CVPR_humanml3d}
  Chuan Guo, Shihao Zou, Xinxin Zuo, Sen Wang, Wei Ji, Xingyu Li, and Li~Cheng.
  \newblock Generating diverse and natural 3d human motions from text.
  \newblock In {\em Proceedings of the IEEE/CVF Conference on Computer Vision and
    Pattern Recognition (CVPR)}, pages 5152--5161, June 2022.
  
  \bibitem{chuan2022tm2t}
  Chuan Guo, Xinxin Zuo, Sen Wang, and Li~Cheng.
  \newblock Tm2t: Stochastic and tokenized modeling for the reciprocal generation
    of 3d human motions and texts.
  \newblock In {\em ECCV}, 2022.
  
  \bibitem{guo2020action2motion}
  Chuan Guo, Xinxin Zuo, Sen Wang, Shihao Zou, Qingyao Sun, Annan Deng, Minglun
    Gong, and Li~Cheng.
  \newblock Action2motion: Conditioned generation of 3d human motions.
  \newblock In {\em Proceedings of the 28th ACM International Conference on
    Multimedia}, pages 2021--2029, 2020.
  
  \bibitem{guzhov2022audioclip}
  Andrey Guzhov, Federico Raue, J{\"o}rn Hees, and Andreas Dengel.
  \newblock Audioclip: Extending clip to image, text and audio.
  \newblock In {\em ICASSP 2022-2022 IEEE International Conference on Acoustics,
    Speech and Signal Processing (ICASSP)}, pages 976--980. IEEE, 2022.
  
  \bibitem{ho2020denoising}
  Jonathan Ho, Ajay Jain, and Pieter Abbeel.
  \newblock Denoising diffusion probabilistic models.
  \newblock {\em Advances in Neural Information Processing Systems},
    33:6840--6851, 2020.
  
  \bibitem{huang2023language}
  Shaohan Huang, Li~Dong, Wenhui Wang, Yaru Hao, Saksham Singhal, Shuming Ma,
    Tengchao Lv, Lei Cui, Owais~Khan Mohammed, Qiang Liu, et~al.
  \newblock Language is not all you need: Aligning perception with language
    models.
  \newblock {\em arXiv preprint arXiv:2302.14045}, 2023.
  
  \bibitem{karras2017audio}
  Tero Karras, Timo Aila, Samuli Laine, Antti Herva, and Jaakko Lehtinen.
  \newblock Audio-driven facial animation by joint end-to-end learning of pose
    and emotion.
  \newblock {\em ACM Transactions on Graphics (TOG)}, 36(4):1--12, 2017.
  
  \bibitem{kim2022flame}
  Jihoon Kim, Jiseob Kim, and Sungjoon Choi.
  \newblock Flame: Free-form language-based motion synthesis \& editing.
  \newblock {\em arXiv preprint arXiv:2209.00349}, 2022.
  
  \bibitem{kudo2018sentencepiece}
  Taku Kudo and John Richardson.
  \newblock Sentencepiece: A simple and language independent subword tokenizer
    and detokenizer for neural text processing.
  \newblock {\em arXiv preprint arXiv:1808.06226}, 2018.
  
  \bibitem{li2023blip2}
  Junnan Li, Dongxu Li, Silvio Savarese, and Steven Hoi.
  \newblock Blip-2: Bootstrapping language-image pre-training with frozen image
    encoders and large language models.
  \newblock {\em arXiv preprint arXiv:2301.12597}, 2023.
  
  \bibitem{li2022blip}
  Junnan Li, Dongxu Li, Caiming Xiong, and Steven Hoi.
  \newblock Blip: Bootstrapping language-image pre-training for unified
    vision-language understanding and generation.
  \newblock In {\em International Conference on Machine Learning}, pages
    12888--12900. PMLR, 2022.
  
  \bibitem{li2021piano}
  Yuwei Li, Minye Wu, Yuyao Zhang, Lan Xu, and Jingyi Yu.
  \newblock Piano: A parametric hand bone model from magnetic resonance imaging.
  \newblock {\em arXiv preprint arXiv:2106.10893}, 2021.
  
  \bibitem{li2022nimble}
  Yuwei Li, Longwen Zhang, Zesong Qiu, Yingwenqi Jiang, Nianyi Li, Yuexin Ma,
    Yuyao Zhang, Lan Xu, and Jingyi Yu.
  \newblock Nimble: a non-rigid hand model with bones and muscles.
  \newblock {\em ACM Transactions on Graphics (TOG)}, 41(4):1--16, 2022.
  
  \bibitem{lin2004rouge}
  Chin-Yew Lin.
  \newblock Rouge: A package for automatic evaluation of summaries.
  \newblock In {\em Text summarization branches out}, pages 74--81, 2004.
  
  \bibitem{ma2022multi}
  Hengbo Ma, Jiachen Li, Ramtin Hosseini, Masayoshi Tomizuka, and Chiho Choi.
  \newblock Multi-objective diverse human motion prediction with knowledge
    distillation.
  \newblock In {\em Proceedings of the IEEE/CVF Conference on Computer Vision and
    Pattern Recognition}, pages 8161--8171, 2022.
  
  \bibitem{AMASS_ICCV2019}
  Naureen Mahmood, Nima Ghorbani, Nikolaus~F. Troje, Gerard Pons-Moll, and
    Michael~J. Black.
  \newblock Amass: Archive of motion capture as surface shapes.
  \newblock In {\em Proceedings of the IEEE/CVF International Conference on
    Computer Vision (ICCV)}, October 2019.
  
  \bibitem{mohammad2022clipmesh}
  Nasir Mohammad~Khalid, Tianhao Xie, Eugene Belilovsky, and Tiberiu Popa.
  \newblock Clip-mesh: Generating textured meshes from text using pretrained
    image-text models.
  \newblock In {\em SIGGRAPH Asia 2022 Conference Papers}, pages 1--8, 2022.
  
  \bibitem{ouyang2022instructgpt}
  Long Ouyang, Jeffrey Wu, Xu~Jiang, Diogo Almeida, Carroll Wainwright, Pamela
    Mishkin, Chong Zhang, Sandhini Agarwal, Katarina Slama, Alex Ray, et~al.
  \newblock Training language models to follow instructions with human feedback.
  \newblock {\em Advances in Neural Information Processing Systems},
    35:27730--27744, 2022.
  
  \bibitem{papineni2002bleu}
  Kishore Papineni, Salim Roukos, Todd Ward, and Wei-Jing Zhu.
  \newblock Bleu: a method for automatic evaluation of machine translation.
  \newblock In {\em Proceedings of the 40th annual meeting of the Association for
    Computational Linguistics}, pages 311--318, 2002.
  
  \bibitem{vposer_SMPL-X:2019}
  Georgios Pavlakos, Vasileios Choutas, Nima Ghorbani, Timo Bolkart, Ahmed A.~A.
    Osman, Dimitrios Tzionas, and Michael~J. Black.
  \newblock Expressive body capture: 3d hands, face, and body from a single
    image.
  \newblock In {\em Proceedings IEEE Conf. on Computer Vision and Pattern
    Recognition (CVPR)}, 2019.
  
  \bibitem{petrovich21actor}
  Mathis Petrovich, Michael~J. Black, and G{\"u}l Varol.
  \newblock Action-conditioned 3{D} human motion synthesis with transformer
    {VAE}.
  \newblock In {\em International Conference on Computer Vision (ICCV)}, 2021.
  
  \bibitem{petrovich22temos}
  Mathis Petrovich, Michael~J. Black, and G{\"u}l Varol.
  \newblock {TEMOS}: Generating diverse human motions from textual descriptions.
  \newblock In {\em European Conference on Computer Vision ({ECCV})}, 2022.
  
  \bibitem{Plappert2016kit}
  Matthias Plappert, Christian Mandery, and Tamim Asfour.
  \newblock The kit motion-language dataset.
  \newblock {\em Big Data}, 4(4):236--252, dec 2016.
  
  \bibitem{plappert2018learning}
  Matthias Plappert, Christian Mandery, and Tamim Asfour.
  \newblock Learning a bidirectional mapping between human whole-body motion and
    natural language using deep recurrent neural networks.
  \newblock {\em Robotics and Autonomous Systems}, 109:13--26, 2018.
  
  \bibitem{radford2021clip}
  Alec Radford, Jong~Wook Kim, Chris Hallacy, Aditya Ramesh, Gabriel Goh,
    Sandhini Agarwal, Girish Sastry, Amanda Askell, Pamela Mishkin, Jack Clark,
    et~al.
  \newblock Learning transferable visual models from natural language
    supervision.
  \newblock In {\em International Conference on Machine Learning}, pages
    8748--8763. PMLR, 2021.
  
  \bibitem{radford2018gpt}
  Alec Radford, Karthik Narasimhan, Tim Salimans, Ilya Sutskever, et~al.
  \newblock Improving language understanding by generative pre-training.
  \newblock 2018.
  
  \bibitem{radford2019gpt2}
  Alec Radford, Jeffrey Wu, Rewon Child, David Luan, Dario Amodei, Ilya
    Sutskever, et~al.
  \newblock Language models are unsupervised multitask learners.
  \newblock {\em OpenAI blog}, 1(8):9, 2019.
  
  \bibitem{raffel2020t5}
  Colin Raffel, Noam Shazeer, Adam Roberts, Katherine Lee, Sharan Narang, Michael
    Matena, Yanqi Zhou, Wei Li, and Peter~J Liu.
  \newblock Exploring the limits of transfer learning with a unified text-to-text
    transformer.
  \newblock {\em The Journal of Machine Learning Research}, 21(1):5485--5551,
    2020.
  
  \bibitem{razavi2019vqvae2}
  Ali Razavi, Aaron Van~den Oord, and Oriol Vinyals.
  \newblock Generating diverse high-fidelity images with vq-vae-2.
  \newblock {\em Advances in neural information processing systems}, 32, 2019.
  
  \bibitem{stable_diffusion}
  Robin Rombach, Andreas Blattmann, Dominik Lorenz, Patrick Esser, and BjÃ¶rn
    Ommer.
  \newblock High-resolution image synthesis with latent diffusion models.
  \newblock In {\em Proceedings of the IEEE Conference on Computer Vision and
    Pattern Recognition (CVPR)}, 2022.
  
  \bibitem{romero2022embodied}
  Javier Romero, Dimitrios Tzionas, and Michael~J Black.
  \newblock Embodied hands: Modeling and capturing hands and bodies together.
  \newblock {\em arXiv preprint arXiv:2201.02610}, 2022.
  
  \bibitem{Rueeg:CVPR:2022}
  Nadine Rueegg, Silvia Zuffi, Konrad Schindler, and Michael~J. Black.
  \newblock {BARC}: Learning to regress {3D} dog shape from images by exploiting
    breed information.
  \newblock In {\em IEEE/CVF Conf.~on Computer Vision and Pattern Recognition
    (CVPR)}, pages 3876--3884, June 2022.
  
  \bibitem{shafir2023priormdm}
  Yonatan Shafir, Guy Tevet, Roy Kapon, and Amit~H Bermano.
  \newblock Human motion diffusion as a generative prior.
  \newblock {\em arXiv preprint arXiv:2303.01418}, 2023.
  
  \bibitem{siyao2022bailando}
  Li~Siyao, Weijiang Yu, Tianpei Gu, Chunze Lin, Quan Wang, Chen Qian,
    Chen~Change Loy, and Ziwei Liu.
  \newblock Bailando: 3d dance generation by actor-critic gpt with choreographic
    memory.
  \newblock In {\em Proceedings of the IEEE/CVF Conference on Computer Vision and
    Pattern Recognition}, pages 11050--11059, 2022.
  
  \bibitem{song2020denoising}
  Jiaming Song, Chenlin Meng, and Stefano Ermon.
  \newblock Denoising diffusion implicit models.
  \newblock {\em arXiv preprint arXiv:2010.02502}, 2020.
  
  \bibitem{takano2015statistical}
  Wataru Takano and Yoshihiko Nakamura.
  \newblock Statistical mutual conversion between whole body motion primitives
    and linguistic sentences for human motions.
  \newblock {\em The International Journal of Robotics Research},
    34(10):1314--1328, 2015.
  
  \bibitem{tevet2022motionclip}
  Guy Tevet, Brian Gordon, Amir Hertz, Amit~H Bermano, and Daniel Cohen-Or.
  \newblock Motionclip: Exposing human motion generation to clip space.
  \newblock In {\em Computer Vision--ECCV 2022: 17th European Conference, Tel
    Aviv, Israel, October 23--27, 2022, Proceedings, Part XXII}, pages 358--374.
    Springer, 2022.
  
  \bibitem{mdm2022human}
  Guy Tevet, Sigal Raab, Brian Gordon, Yonatan Shafir, Amit~H Bermano, and Daniel
    Cohen-Or.
  \newblock Human motion diffusion model.
  \newblock {\em arXiv preprint arXiv:2209.14916}, 2022.
  
  \bibitem{touvron2023llama}
  Hugo Touvron, Thibaut Lavril, Gautier Izacard, Xavier Martinet, Marie-Anne
    Lachaux, Timoth{\'e}e Lacroix, Baptiste Rozi{\`e}re, Naman Goyal, Eric
    Hambro, Faisal Azhar, et~al.
  \newblock Llama: Open and efficient foundation language models.
  \newblock {\em arXiv preprint arXiv:2302.13971}, 2023.
  
  \bibitem{van2017vqvae}
  Aaron Van Den~Oord, Oriol Vinyals, et~al.
  \newblock Neural discrete representation learning.
  \newblock {\em Advances in neural information processing systems}, 30, 2017.
  
  \bibitem{vedantam2015cider}
  Ramakrishna Vedantam, C~Lawrence~Zitnick, and Devi Parikh.
  \newblock Cider: Consensus-based image description evaluation.
  \newblock In {\em Proceedings of the IEEE conference on computer vision and
    pattern recognition}, pages 4566--4575, 2015.
  
  \bibitem{wang2022beit}
  Wenhui Wang, Hangbo Bao, Li~Dong, Johan Bjorck, Zhiliang Peng, Qiang Liu, Kriti
    Aggarwal, Owais~Khan Mohammed, Saksham Singhal, Subhojit Som, et~al.
  \newblock Image as a foreign language: Beit pretraining for all vision and
    vision-language tasks.
  \newblock {\em arXiv preprint arXiv:2208.10442}, 2022.
  
  \bibitem{wei2021finetuned}
  Jason Wei, Maarten Bosma, Vincent~Y Zhao, Kelvin Guu, Adams~Wei Yu, Brian
    Lester, Nan Du, Andrew~M Dai, and Quoc~V Le.
  \newblock Finetuned language models are zero-shot learners.
  \newblock {\em arXiv preprint arXiv:2109.01652}, 2021.
  
  \bibitem{chen2023mld}
  Chen Xin, Biao Jiang, Wen Liu, Zilong Huang, Bin Fu, Tao Chen, Jingyi Yu, and
    Gang Yu.
  \newblock Executing your commands via motion diffusion in latent space.
  \newblock In {\em Proceedings of the IEEE/CVF Conference on Computer Vision and
    Pattern Recognition (CVPR)}, June 2023.
  
  \bibitem{xu2021videoclip}
  Hu~Xu, Gargi Ghosh, Po-Yao Huang, Dmytro Okhonko, Armen Aghajanyan, Florian
    Metze, Luke Zettlemoyer, and Christoph Feichtenhofer.
  \newblock Videoclip: Contrastive pre-training for zero-shot video-text
    understanding.
  \newblock {\em arXiv preprint arXiv:2109.14084}, 2021.
  
  \bibitem{yamada2018rae}
  Tatsuro Yamada, Hiroyuki Matsunaga, and Tetsuya Ogata.
  \newblock Paired recurrent autoencoders for bidirectional translation between
    robot actions and linguistic descriptions.
  \newblock {\em IEEE Robotics and Automation Letters}, 3(4):3441--3448, 2018.
  
  \bibitem{youwang2022clipactor}
  Kim Youwang, Kim Ji-Yeon, and Tae-Hyun Oh.
  \newblock Clip-actor: Text-driven recommendation and stylization for animating
    human meshes.
  \newblock In {\em Computer Vision--ECCV 2022: 17th European Conference, Tel
    Aviv, Israel, October 23--27, 2022, Proceedings, Part III}, pages 173--191.
    Springer, 2022.
  
  \bibitem{yuan2020dlow}
  Ye~Yuan and Kris Kitani.
  \newblock Dlow: Diversifying latent flows for diverse human motion prediction.
  \newblock In {\em Computer Vision--ECCV 2020: 16th European Conference,
    Glasgow, UK, August 23--28, 2020, Proceedings, Part IX 16}, pages 346--364.
    Springer, 2020.
  
  \bibitem{zhang2023generating}
  Jianrong Zhang, Yangsong Zhang, Xiaodong Cun, Shaoli Huang, Yong Zhang, Hongwei
    Zhao, Hongtao Lu, and Xi~Shen.
  \newblock T2m-gpt: Generating human motion from textual descriptions with
    discrete representations.
  \newblock In {\em Proceedings of the IEEE/CVF Conference on Computer Vision and
    Pattern Recognition (CVPR)}, 2023.
  
  \bibitem{zhang2022motiondiffuse}
  Mingyuan Zhang, Zhongang Cai, Liang Pan, Fangzhou Hong, Xinying Guo, Lei Yang,
    and Ziwei Liu.
  \newblock Motiondiffuse: Text-driven human motion generation with diffusion
    model.
  \newblock {\em arXiv preprint arXiv:2208.15001}, 2022.
  
  \bibitem{zhang2022opt}
  Susan Zhang, Stephen Roller, Naman Goyal, Mikel Artetxe, Moya Chen, Shuohui
    Chen, Christopher Dewan, Mona Diab, Xian Li, Xi~Victoria Lin, et~al.
  \newblock Opt: Open pre-trained transformer language models.
  \newblock {\em arXiv preprint arXiv:2205.01068}, 2022.
  
  \bibitem{zhang2019bertscore}
  Tianyi Zhang, Varsha Kishore, Felix Wu, Kilian~Q Weinberger, and Yoav Artzi.
  \newblock Bertscore: Evaluating text generation with bert.
  \newblock {\em arXiv preprint arXiv:1904.09675}, 2019.
  
  \bibitem{zhang2021we}
  Yan Zhang, Michael~J Black, and Siyu Tang.
  \newblock We are more than our joints: Predicting how 3d bodies move.
  \newblock In {\em Proceedings of the IEEE/CVF Conference on Computer Vision and
    Pattern Recognition}, pages 3372--3382, 2021.
  
  \bibitem{Zuffi:CVPR:2018}
  Silvia Zuffi, Angjoo Kanazawa, and Michael~J. Black.
  \newblock Lions and tigers and bears: Capturing non-rigid, {3D}, articulated
    shape from images.
  \newblock In {\em IEEE Conference on Computer Vision and Pattern Recognition
    (CVPR)}, pages 3955--3963. IEEE Computer Society, 2018.
  
  \end{thebibliography}

}

\newpage
\section*{\centering {\LARGE Appendix}}
\vspace{80pt}

\renewcommand\thesection{\Alph{section}}
\renewcommand*{\theHsection}{appedix.\thesection}
\setcounter{section}{0}
\setcounter{figure}{3}
\setcounter{table}{6}
\setcounter{equation}{4}

This appendix provides qualitative comparison results (\cref{sec:appendix:qualitative}), 
additional experiments (\cref{sec:appendix:exps}) on the components of MotionGPT models, inference time (\cref{sec:inferencetime}), statistics on motion vocabulary (\cref{sec:appendix:visual}),
evaluations on hyperparameters (\cref{sec:appendix:hyperparameters}), user study (\cref{sec:appendix:userstudy}), a protocol for the uniform evaluation (\cref{sec:appendix:data: details}), and more implementation details (\cref{sec:appendix:model:details}) of MotionGPT models .
Please note evaluations on our training scheme (\cref{sec:pretrian}), elaborations on the difference of T2M-GPT (\cref{sec:appendix:diff}),
implementation details of motion completion (\cref{sec:appendix:reimplement}),
and more metric definitions (\cref{appedix:metrics:details}).

\textbf{Video.} We have provided the supplemental video to illustrate our results. In this video, we show 1) comparisons of text-to-motion, 2) comparisons of motion captioning, and 3) more results on motion prediction and other tasks. We suggest watching this video for dynamic motion results.

\textbf{Code} is available in supplements. We provide example code files, which include the process of the training and evaluation of our MotionGPT models, as well as several example results. We can hardly upload our large model files, but all codes, data, and pre-trained models will be fully released.



\section{Qualitative Results}
\label{sec:appendix:qualitative}

\begin{figure}[h]
    \centering
    \includegraphics[width=\linewidth]{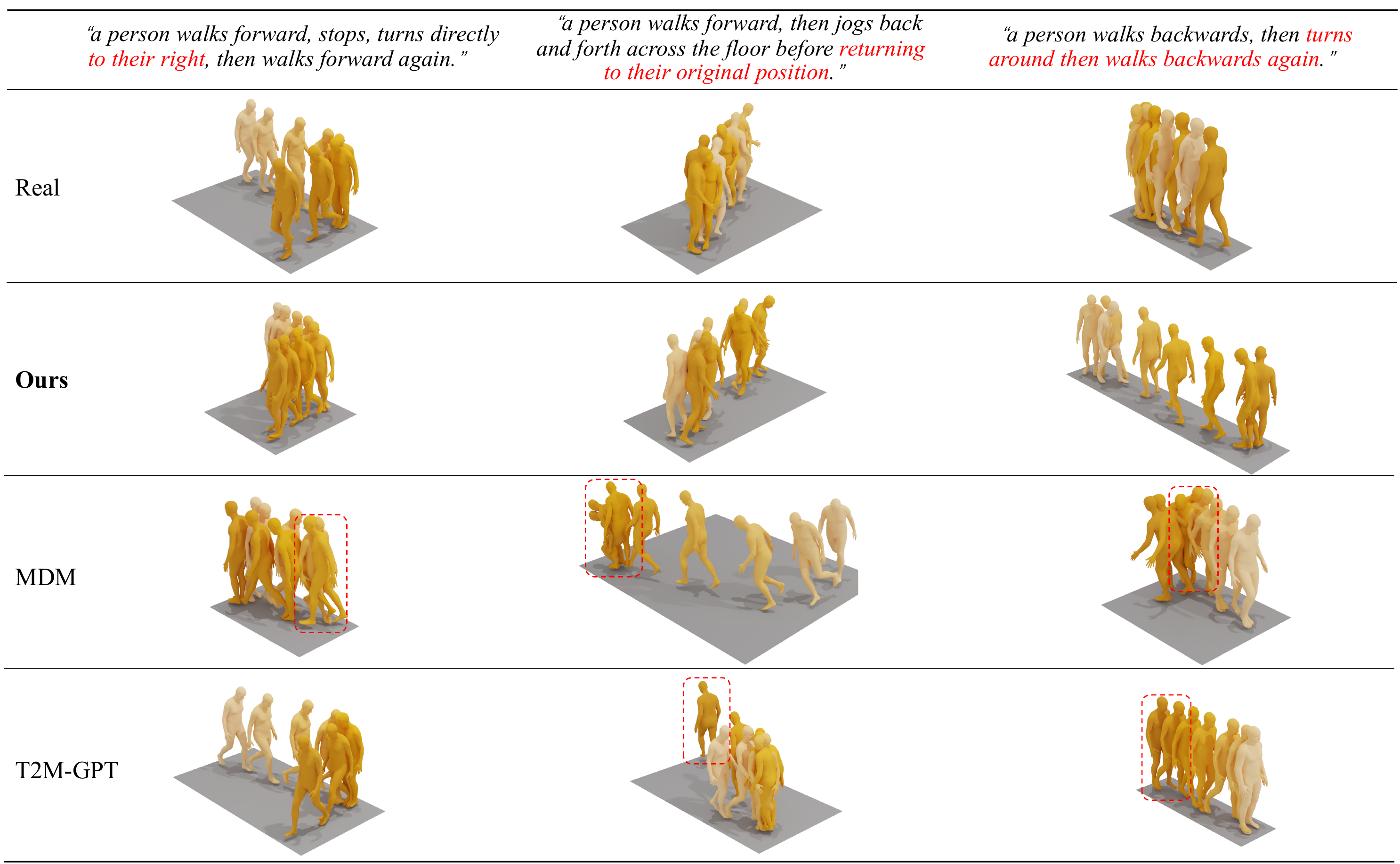}
    \caption{Comparison on text-driven motion generation. The provided state-of-the-art methods are under the same training and inference setting on HumanML3D \cite{Guo_2022_CVPR_humanml3d}. The red words and boxes highlight the misaligned motions. The results demonstrate that our motion-language per-training shows promising text understanding for motion generation.}
    \label{fig:appendix:compt2m}
\end{figure}

\begin{figure}[!htb]
    \centering
    \includegraphics[width=0.99\linewidth]{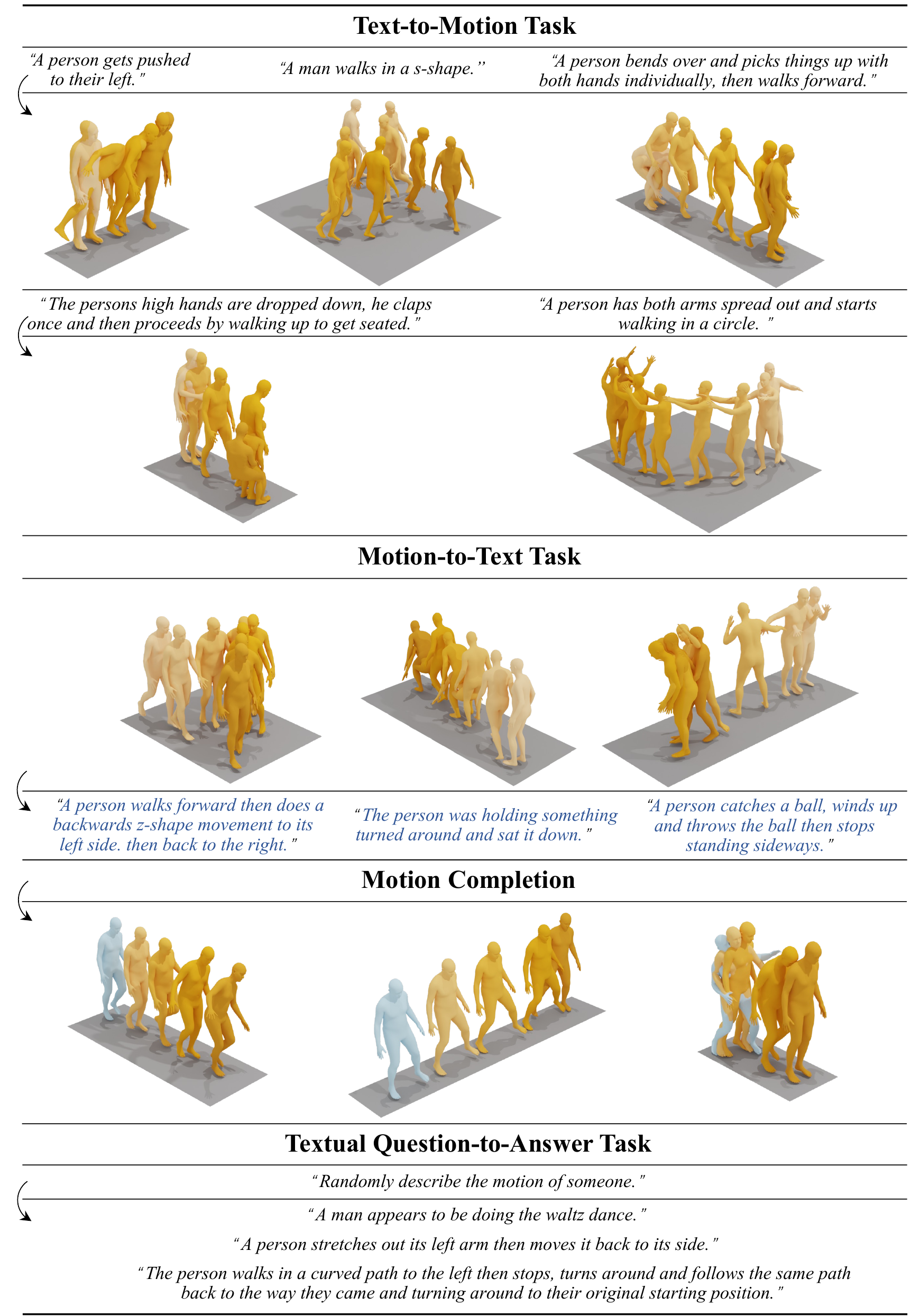}
    \caption{Gallery for the results of our unified MotionGPT. More samples are from our best model for text-to-motion synthesis, motion captioning, and textual question-to-answer task. The supervision of MotionGPT relies on our instruction-based motion-language dataset ($cf.$ \cref{sec:appendix:data: details}) based on previous motion datasets~\cite{Guo_2022_CVPR_humanml3d, AMASS_ICCV2019}. We recommend the dynamic visualization in our supplemental video. }
    \label{fig:appendix:more_rs}
\end{figure}

\begin{figure}[h]
    \centering
    \includegraphics[width=\linewidth]{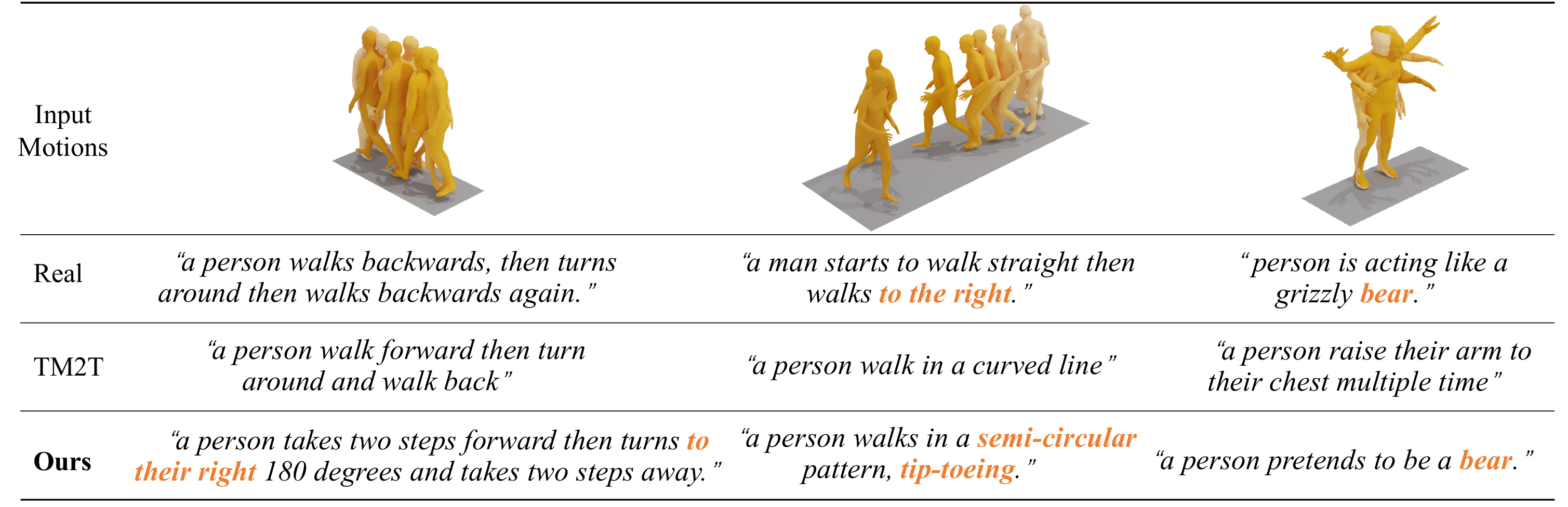}
    \vspace{-10pt}
    \caption{Comparison of the state-of-the-art method on motion captioning task. All provided methods are under the same training and inference setting on HumanML3D~\cite{Guo_2022_CVPR_humanml3d}. The results demonstrate that our text descriptions correspond better to the motion and have correct grammar. The orange words indicate the matching results, while the red marks the incorrect grammar.}
    \label{fig:appendix:compm2t}
\end{figure}

\section{Additional Experiments}
\label{sec:appendix:exps}
We conduct several experiments to continue the evaluations of MotionGPT models. We first evaluate the text-to-motion results on KIT dataset (\cref{sec:t2mkit}). Then we evaluate the hyperparameters of motion tokenizer $\mathcal{V}$ (\cref{sec:vqvae}). After that, we study the effectiveness of the training scheme (\cref{sec:pretrian}). We also provide the elaboration on the difference of T2M-GPT (\cref{sec:appendix:diff}),
implementation details of motion completion (\cref{sec:appendix:reimplement}).

\subsection{Text-to-Motion on KIT dataset.} 
\label{sec:t2mkit}
Following the same procedure on HumanML3D\cite{Guo_2022_CVPR_humanml3d} dataset, We train a 220M MotionGPT base model on the KIT\cite{Plappert2016kit} dataset without any pre-training. We evaluate this model under the same settings of \cite{Guo_2022_CVPR_humanml3d}. Most results are borrowed from their own paper of the benchmark in \cite{Guo_2022_CVPR_humanml3d}. \cref{tab:tm:comp:kit} shows that MotionGPT achieves comparable performance compared to the previous state-of-the-arts.

\begin{table}[h]
\vspace{20pt}
\resizebox{\columnwidth}{!}{%
\begin{tabular}{@{}lcccccccc@{}}
\toprule
\multirow{2}{*}{Methods}&\multicolumn{3}{c}{RPrecision$\uparrow$}&\multicolumn{1}{c}{\multirow{2}{*}{FID$\downarrow$}}&\multirow{2}{*}{MMDist$\downarrow$}&\multirow{2}{*}{Diversity$\rightarrow$}&\multirow{2}{*}{MModality$\uparrow$}\\\cmidrule(lr){2-4}
&\multicolumn{1}{c}{Top1}&\multicolumn{1}{c}{Top2}&\multicolumn{1}{c}{Top3}&\multicolumn{1}{c}{}&&&\\\midrule
Real&
$0.424^{\pm.005}$&
$0.649^{\pm.006}$&
$0.779^{\pm.006}$&
$0.031^{\pm.004}$&
$2.788^{\pm.012}$&
$11.08^{\pm.097}$&
\multicolumn{1}{c}{-}
\\\midrule
TM2T\cite{chuan2022tm2t}&
$0.280^{\pm.005}$&
$0.463^{\pm.006}$&
$0.587^{\pm.005}$&
$3.599^{\pm.153}$&
$4.591^{\pm.026}$&
$9.473^{\pm.117}$&
${3.292}^{\pm.081}$\\
MDM\cite{mdm2022human}&
$0.164^{\pm.004}$&
$0.291^{\pm.004}$&
$0.396^{\pm.004}$&
${0.497}^{\pm.021}$&
$9.191^{\pm.022}$&
${10.85}^{\pm.109}$&
${1.907}^{\pm.214}$\\
MLD\cite{chen2023mld}&
${0.390}^{\pm.008}$&
${0.609}^{\pm.008}$&
${0.734}^{\pm.007}$&
${0.404}^{\pm.027}$&
${3.204}^{\pm.027}$&
$10.80^{\pm.117}$&
${2.192}^{\pm.071}$\\
T2M-GPT \cite{zhang2023generating}    &$0.416^{\pm.006}$ & $0.627^{\pm.006}$ & $0.745^{\pm.006}$ & $0.514^{\pm.029}$ & $3.007^{\pm.023}$ & $10.92^{\pm.108}$ & $1.570^{\pm.039}$  \\
\midrule
MotionGPT (Ours) &
$0.366^{\pm.005}$ & $0.558^{\pm.004}$ & $0.680^{\pm.005}$ & $0.510^{\pm.016}$ & $3.527^{\pm.021}$ & $10.35^{\pm.084}$ & 
$2.328^{\pm.117}$
\\\bottomrule
\end{tabular}%
}
\vspace{8pt}
\caption{We involve KIT~\cite{Plappert2016kit}dataset and evaluate the methods on the text-driven motion generation task. Please refer to Tab. 3 for more details on metrics and notations.
}
\label{tab:tm:comp:kit}
\end{table}
     
\subsection{Ablation on Motion Tokenizer.} 
\label{sec:vqvae}
We ablate the motion tokenizer $\mathcal{V}$ of our MotionGPT models, studying the size $K$ of motion codebooks. We also compare this VQ-VAE with other VAE models in previous works~\cite{vposer_SMPL-X:2019, petrovich21actor, chen2023mld}, as shown in \cref{tab:mr:ablation}. This comparison demonstrates the improvement of VQ-VAE on motion reconstruction. With this ablation studies on the codebook size $K$, we thus select $K=512$ for most experiments.

\begin{table}[t]
\centering
\resizebox{0.7\columnwidth}{!}{%
\begin{tabular}{@{}lccccc@{}}
\toprule
\multirow{2}{*}{Method}&\multicolumn{5}{c}{Reconstruction}
\\\cmidrule(lr){2-6}
&MPJPE$\downarrow$&PAMPJPE$\downarrow$&ACCL$\downarrow$&FID$\downarrow$&DIV$\rightarrow$
\\ \midrule
Real  &  \multicolumn{1}{c}{-} & \multicolumn{1}{c}{-}& \multicolumn{1}{c}{-} &  $0.002\hidden{^{\pm.000}}$&	$9.503\hidden{^{\pm.065}}$     \\ \midrule
VPoser-t~\cite{vposer_SMPL-X:2019}  & $75.6$ & $48.6$& $9.3$ &     $1.430\hidden{^{\pm.002}}$ & $8.336\hidden{^{\pm.099}}$   \\
ACTOR~\cite{petrovich21actor}     & $65.3$ & $41.0$& $\boldsymbol{7.0}$ &    $0.341\hidden{^{\pm.002}}$ & $\boldsymbol{9.569}\hidden{^{\pm.099}}$ \\
MLD-1~\cite{chen2023mld}  & $\boldsymbol{54.4}$ & $41.6$ & $8.3$ & $0.247\hidden{^{\pm.001}}$ & $9.630\hidden{^{\pm.081}}$ \\
 \midrule
MotionGPT (Ours) & 
$55.8$ &$\boldsymbol{40.1}$  & $7.5$	&$\boldsymbol{0.067}\hidden{^{\pm.001}}$&$9.675\hidden{^{\pm.064}}$	
\\ \bottomrule
\toprule
$K=256$ & $76.4$ & $51.3$ & $10.0$ & $0.187\hidden{^{\pm.002}}$ & $\boldsymbol{9.496}\hidden{^{\pm.070}}$ \\ 
$K=512$ & $\boldsymbol{55.8}$ &$\boldsymbol{40.1}$  & $\boldsymbol{7.5}$	&$\boldsymbol{0.067}\hidden{^{\pm.001}}$&$9.675\hidden{^{\pm.064}}$ \\
$K=1024$ & $60.3$ & $44.0$ & $8.6$ & $0.086\hidden{^{\pm.001}}$ & $9.677\hidden{^{\pm.078}}$\\
$K=2048$ & $78.9$ & $51.4$ & $10.5$ & $0.141\hidden{^{\pm.001}}$ & $9.597\hidden{^{\pm.056}}$
\\ \bottomrule
\end{tabular}%
}
\vspace{10pt}
\caption{Evaluation of our motion tokenizer on the motion part of HumanML3D~\cite{Guo_2022_CVPR_humanml3d} dataset. We follow MLD~\cite{chen2023mld} to evaluate our VQ-VAE model $\mathcal{V}$: MPJPE and PAMPJPE are measured in millimeter. ACCL indicates acceleration error. We evaluate FID and Diversity the same as Tab. 3. The baselines of VPoser-t~\cite{vposer_SMPL-X:2019} and ACTOR~\cite{petrovich21actor} are borrowed from MLD. $K$ indicates the codebook size, and $K=512$ shows the best performance of motion reconstruction.}
\label{tab:mr:ablation}
\end{table}

\newpage
\label{sec:pretrian}
\subsection{Effectiveness of Training Scheme} 
\textbf{Motion-Language Pre-training} vs \textbf{Instructions Tuning}. We have provided the illustration of our training scheme in Fig. 3 and the evaluation in Tab. 6. We further ablate this training scheme on the base MotionGPT model, by evaluating the motion-language pre-training (the second step) and instruction tuning (the third step). As shown in \cref{tab:abl:pretrain}, we train these models with the same 600K iterations. Compared to other training combinations, the full-stage MotionGPT achieves higher performance on most motion tasks.

\begin{table}[h]
\resizebox{\columnwidth}{!}{%
\begin{tabular}{@{}lcccccccccccc@{}}
\toprule
\multirow{2}{*}{Size} & 
\multirow{2}{*}{Pre-training} & 
\multirow{2}{*}{Instruction Tuning} & 
\multicolumn{3}{c}{Text-to-Motion}& 
\multicolumn{3}{c}{Motion-to-Text} & 
\multicolumn{2}{c}{Motion Prediction}&
\multicolumn{2}{c}{Motion In-between}
\\ \cmidrule(lr){4-6} \cmidrule(lr){7-9} 
\cmidrule(lr){10-11} 
\cmidrule(lr){12-13}
&&& R TOP3 $\uparrow$ & FID $\downarrow$ & DIV $\rightarrow$ & MMDist$\downarrow$ & Bleu@4$\uparrow$ & Cider$\uparrow$ 
&   FID $\downarrow$ & DIV $\rightarrow$  
& FID $\downarrow$ & DIV $\rightarrow$ 
\\ \midrule
Real & - & - &
$0.797\hidden{^{\pm.002}}$ &
$0.002\hidden{^{\pm.000}}$ &
$9.503\hidden{^{\pm.065}}$ &
${2.901}$
& -& -
& $0.002$ & $9.503$ 
& $0.002$ & $9.503$ 
\\ \midrule
Base & \CheckmarkBold & \XSolidBrush&
$\boldsymbol{0.722}\hidden{^{\pm.002}}$	& $0.365\hidden{^{\pm.008}}$	& 
$9.407\hidden{^{\pm.042}}$ &
$\boldsymbol{2.821}$	& $\boldsymbol{12.47}$ & $\boldsymbol{29.2}$ 
& - & - 
& - & - 
\\
Base & \CheckmarkBold &  \CheckmarkBold &
$0.700$	& $\boldsymbol{0.160}$	& $\boldsymbol{9.411}$ &
$3.019$	& $11.42$ & $28.2$ 
& ${0.905}$ &${8.972}$
&$\boldsymbol{0.214}$&$\boldsymbol{9.560}$
\\
Base & \XSolidBrush  & \CheckmarkBold & 0.607 & 0.324 & 9.563 & 3.374 & $10.92$& $27.7$&  $1.643$ & $8.829$ & $0.323$  & $9.628$
\\ 
\bottomrule
\end{tabular}%
}
\vspace{10pt}
\caption{Evaluation of the training scheme on the base MotionGPT models. We evaluate the results with the proposed evaluation protocols in \cref{sec:appendix:data: details}. Please refer to Tab. 2 for metrics and the details.}
\label{tab:abl:pretrain}
\end{table}

\label{sec:finetuning}
\textbf{Instructions Tuning} vs \textbf{Task-Specific Tuning}. While our unified instruction-tuned MotionGPT model has demonstrated competitive performance across various motion-related tasks, further fine-tuning can always enhance its performance on specific tasks. Therefore, we focus on the text-to-motion task and motion in-between task as illustrative examples to showcase the performance of the model before and after fine-tuning. By comparing the results in \cref{tab:tm:abl:finetune}, we can assess the effectiveness of fine-tuning in improving task-specific performance.

\begin{table}[h]
\centering
\resizebox{0.8\columnwidth}{!}{%
\begin{tabular}{@{}cccccccccc@{}}
\toprule
\multirow{2}{*}{Insturct tuned} & 
\multirow{2}{*}{Fine tuned} & & 
\multicolumn{3}{c}{Text-to-Motion}& & 
\multicolumn{3}{c}{Motion In-between}
\\ \cmidrule(lr){4-6}  \cmidrule(lr){8-10} 
&& 
& R TOP1$\uparrow$ 
& FID$\downarrow$ 
& DIV$\rightarrow$ 
&&$\text{FID}\downarrow $  
& DIV$\uparrow$            
& ADE$\downarrow$
\\ \midrule
\CheckmarkBold & \XSolidBrush & &
${0.435}\hidden{^{\pm.003}}$ & $\boldsymbol{0.160}\hidden{^{\pm.008}}$ & $9.411\hidden{^{\pm.081}}$ 
&&
${0.214}$&$\boldsymbol{9.560}$&${3.762}$
\\
\CheckmarkBold & \CheckmarkBold  & &
$\boldsymbol{0.492}\hidden{^{\pm.003}}$ &
${0.232}\hidden{^{\pm.008}}$ & 
$\boldsymbol{9.528}\hidden{^{\pm.071}}$ &
&$\boldsymbol{0.209}$&$9.378$&$\boldsymbol{3.281}$
\\ \bottomrule
\end{tabular}%
}
\vspace{10pt}
\caption{Evaluation of new task tuning of different size models on HumanML3D~\cite{Guo_2022_CVPR_humanml3d} dataset. 
}
\label{tab:tm:abl:finetune}
\end{table}

\subsection{Difference of T2M-GPT}
\label{sec:appendix:diff}
We introduce the difference between T2M-GPT~\cite{zhang2023generating} to show our unified framework. T2M-GPT investigates a generative framework based on VQ-VAE and Transformer for motion generation only. 
They incorporate language information by leveraging CLIP \cite{radford2021clip} to extract text embedding as motion generation conditions, which is similar to most previous work, such as MDM~\cite{mdm2022human}, MLD~\cite{chen2023mld}, and MotionDiffuse~\cite{zhang2022motiondiffuse}.
%
%
However, our MotionGPTs are based on the pre-trained language model so it naturally leverages the strong language generation and zero-shot transfer abilities of pre-trained language models. Benefiting from the motion-language vocabulary, MotionGPT thus generates both human language and human motion in a unified model.

\subsection{Implementation details of Motion Completion}
\label{sec:appendix:reimplement}
Please note that MDM\cite{mdm2022human} accomplish motion in-between task in their paper through masked motion “in-painting” which fix the first and last 25\% of the motion, leaving the model to generate the remaining 50\% in the middle. To achieve the motion prediction task with MDM, we fix the first 20\% of the motion and then generate the remaining. All our results are computed by utilizing their provided pre-trained model. To compare with MDM in Tab. 5 on both motion in-between and motion prediction tasks, we evaluate our MotionGPT with the same setting during the inference.

\section{Inference Time}
\label{sec:inferencetime}
We provide a detailed study on inference time with our different model sizes below. Due to our auto-regressive model for motion generation, we use Frames Per Second (FPS) to evaluate our time costs. All the time costs are evaluated on 8 Tesla V100 using one batch size. \cref{tab:tm:abl:scheduler} shows that any size of our MotionGPTs can support real-time human animations and come up to hundreds of FPS.

\begin{table}[h]
\centering
\resizebox{0.45\columnwidth}{!}
{%
\begin{tabular}{@{}llcc@{}}
\toprule
Models & Backbone&Parameters
& FPS $\uparrow$
\\ \midrule
MotionGPT & Small & 60 M 
& 421.31
\\
MotionGPT &Base & 220 M
& 222.69
\\
MotionGPT &Large & 770 M
& 119.75
\\
\bottomrule
\end{tabular}%
}
\vspace{5pt}
\caption{Evaluation of inference time costs on text-driven motion generation. We evaluate the Frames Per Second (FPS) by averaging our generated frames for each second. We show the time costs on different model sizes. Under the same 1 Tesla V100, the smaller model size gets the faster FPS. All models can support real-time motion animation applications.}
\vspace{-5pt}
\label{tab:tm:abl:scheduler}
\end{table}

\section{Statistics on Motion Vocabulary}
\label{sec:appendix:visual}
\begin{figure}[!h]
\centering
\begin{minipage}[t]{0.32\textwidth}
\centering
\includegraphics[width=5cm]{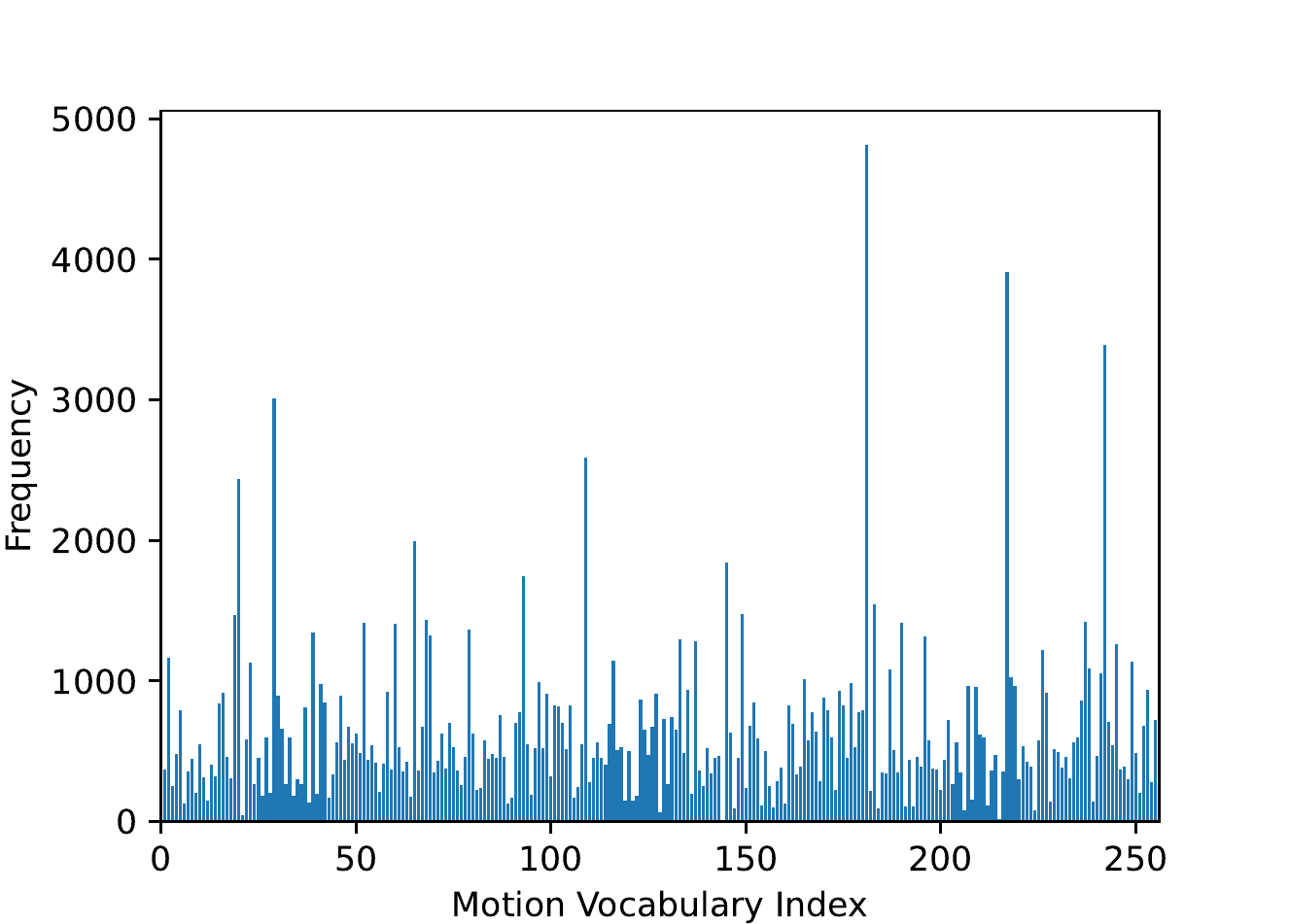}
\end{minipage}
\begin{minipage}[t]{0.32\textwidth}
\centering
\includegraphics[width=5cm]{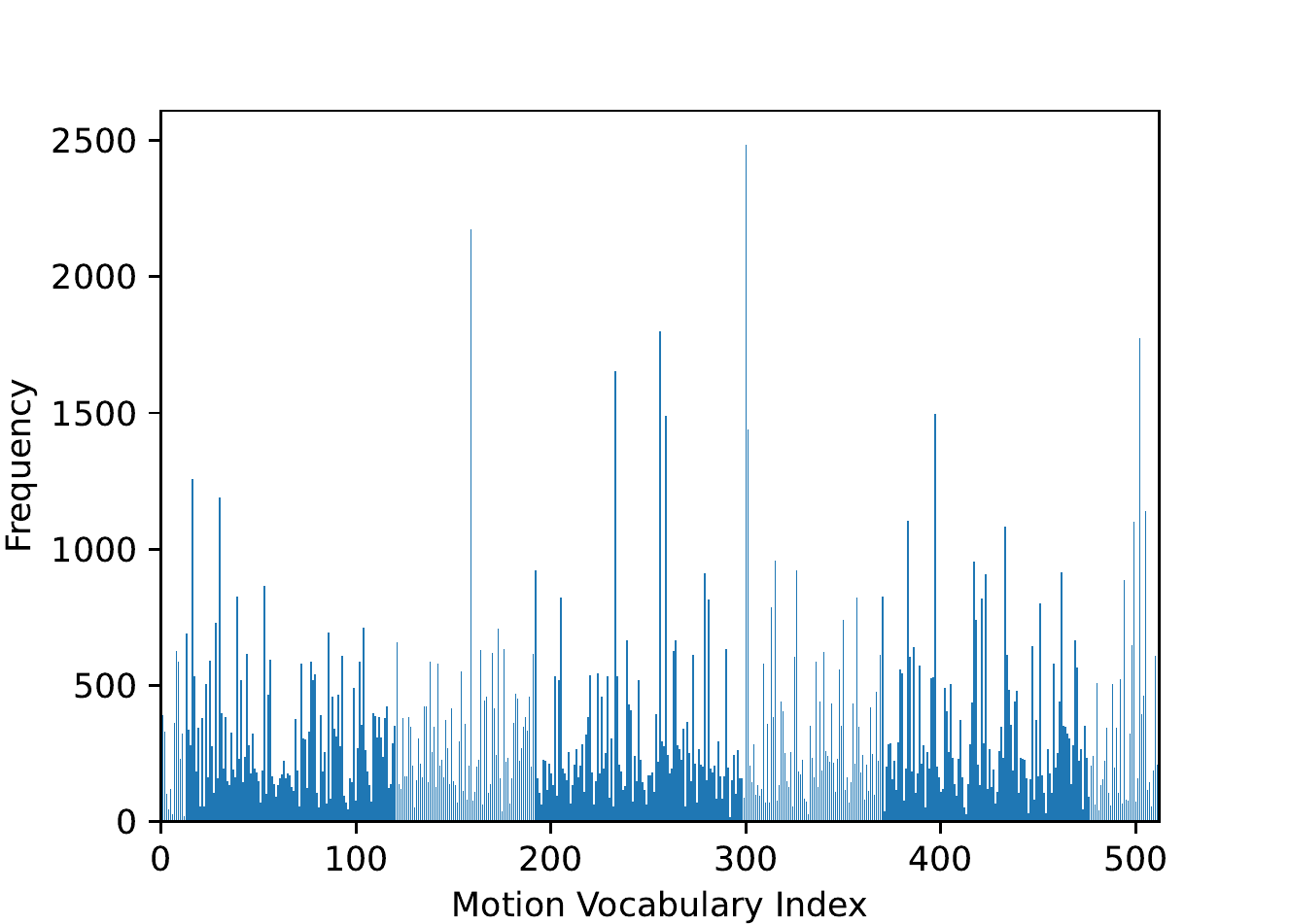}
\end{minipage}
\begin{minipage}[t]{0.32\textwidth}
\centering
\includegraphics[width=5cm]{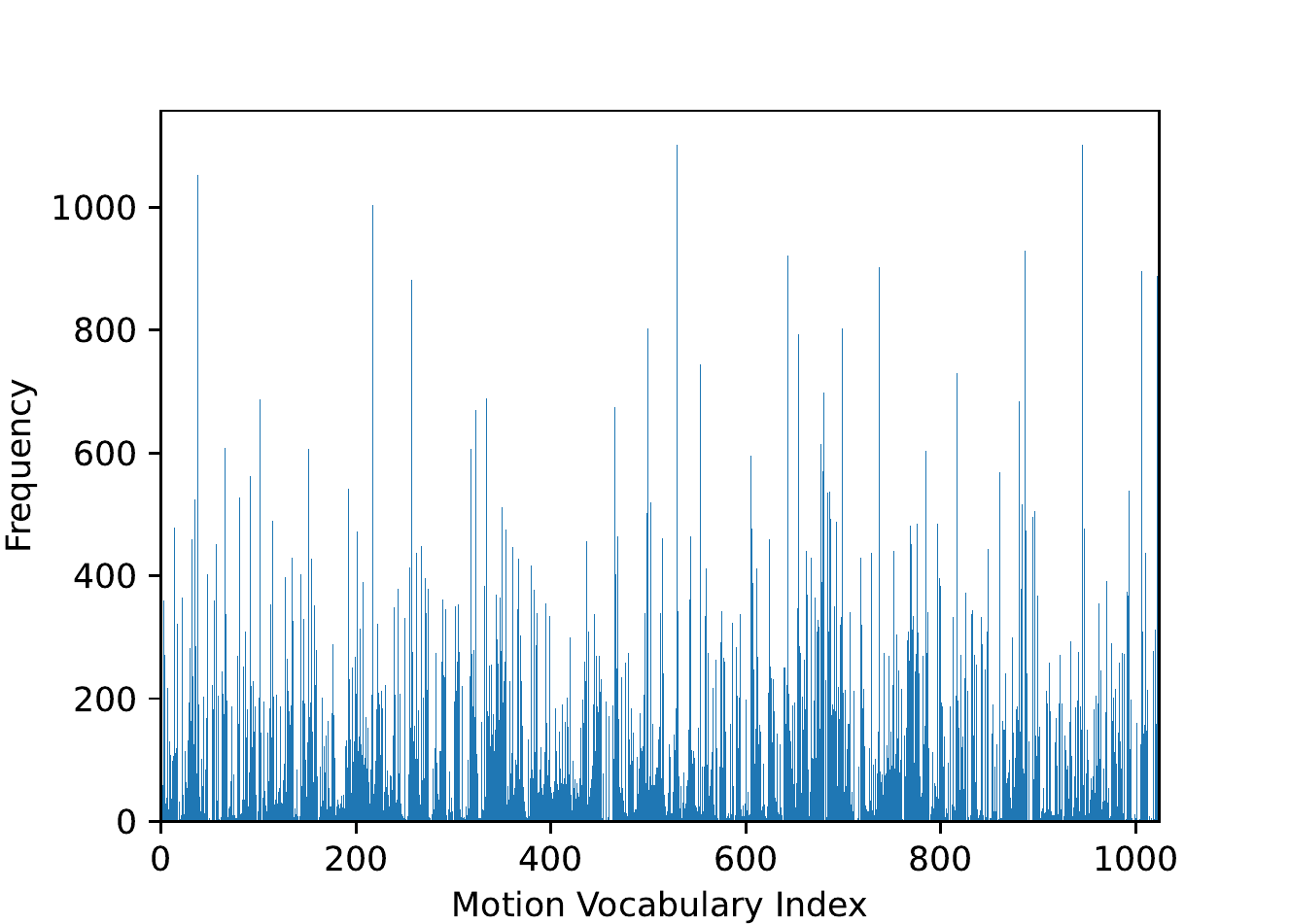}
\end{minipage}
\caption{The statistics of each "word" in different sizes of motion vocabulary $V_m$. From left to right, the vocabulary size is $K=256, 512,1 024$. ($cf$. \cref{tab:mr:ablation}, $K=512$ for the best motion quality.) }
\label{fig:codebook}
\end{figure}

We visualize the usage of each "word" in our motion vocabulary $V_m$ item generated by our motion tokenizer $\mathcal{V}$. We sample all motions from the whole test set of HumanML3D dataset~\cite{Guo_2022_CVPR_humanml3d} and count each "word". In \cref{fig:codebook}, it shows the utilization of our motion codebook, which seems to be a concise but informative motion vocabulary.

\section{Evaluation of Hyperparameters}
\label{sec:appendix:hyperparameters}
We conduct experiments to investigate the impact of different sampling strategies on the generation results. Specifically, we compare the use of greedy search, which selects the most probable token at each step, with sampling from the probability distribution and adopting beam search, which is evaluated in previous language models~\cite{raffel2020t5}. Beam search expands the search space for improved sequence probability matching. The results in \cref{tab:appendix:guidance} demonstrate that while avoiding sampling and using beam search can slightly improve generation quality, they also significantly reduce the diversity of generated motions from the same text description.

\label{appendix:guidance}

\begin{table}[H]
\centering
\resizebox{\columnwidth}{!}{%
\begin{tabular}{@{}lcccccccc@{}}
\toprule
 Method & \multirow{2}{*}{Sample} & \multirow{2}{*}{$\text{\#beams}$}  & \multicolumn{1}{c}{R Precision} & \multicolumn{1}{c}{\multirow{2}{*}{FID$\downarrow$}} & \multirow{2}{*}{MM Dist$\downarrow$} & \multirow{2}{*}{Diversity$\rightarrow$} & \multirow{2}{*}{MModality$\uparrow$} \\\ &&&  \multicolumn{1}{c}{Top 3$\uparrow$}       & \multicolumn{1}{c}{}                     &                          &                            &     \\ \toprule
Real & - &- &
  $0.797^{\pm.002}$ &
  $0.002^{\pm.000}$ &
  $2.974^{\pm.008}$ &
  $9.503^{\pm.065}$ &
  \multicolumn{1}{c}{-}
  \\ \midrule
\multirow{4}{*}{MotionGPT}  && - & $0.780^{\pm.002}$ & $0.224^{\pm.009}$ & $3.076^{\pm.009}$ & $9.492^{\pm.056}$ & \multicolumn{1}{c}{-}
\\
&&2 & $0.780^{\pm.002}$ & $0.199^{\pm.008}$ & $3.083^{\pm.007}$ & $9.512^{\pm.063}$ & \multicolumn{1}{c}{-}
\\
&&3 & $0.781^{\pm.002}$ & $0.179^{\pm.008}$ & $3.099^{\pm.009}$ & $9.516^{\pm.064}$ &  \multicolumn{1}{c}{-}
\\
&&4 & $0.782^{\pm.002}$ &	$0.160^{\pm.007}$ &	$3.092^{\pm.010}$ &	$9.536^{\pm.060}$	&  \multicolumn{1}{c}{-}

\\ \midrule
\multirow{4}{*}{MotionGPT} &\checkmark & - & ${0.778}^{\pm.002}$ & $0.232^{\pm.008}$ & ${3.096}^{\pm.008}$ & ${9.528}^{\pm.071}$ & ${2.008}^{\pm.084}$
\\
&\checkmark & 2 & $0.780^{\pm.002}$ & $0.194^{\pm.008}$ & $3.091^{\pm.010}$ & $9.508^{\pm.063}$ & {$1.140^{\pm.064}$}
\\
&\checkmark & 3 & $0.780^{\pm.002}$ & $0.190^{\pm.008}$ & $3.089^{\pm.011}$ & $9.529^{\pm.061}$ & $0.929^{\pm.055}$
\\
&\checkmark & 4 & $0.780^{\pm.002}$ & $0.182^{\pm.008}$ & $3.093^{\pm.008}$ & $9.537^{\pm.059}$ & $0.803^{\pm.044}$
\\
\bottomrule
\end{tabular}%
}
\vspace{5pt}
\caption{Evaluations on hyperparameters for MotionGPT generations. We study the influence of two hyperparameters: \textit{sample} stands for sampling from distribution; $\textit{\#beams}$ means the number of beams for beam search, where empty means no beam search.}
\label{tab:appendix:guidance}
\end{table}

\section{User Study}
\label{sec:appendix:userstudy}
\begin{figure}[!h]
\centering
\includegraphics[width=\textwidth]{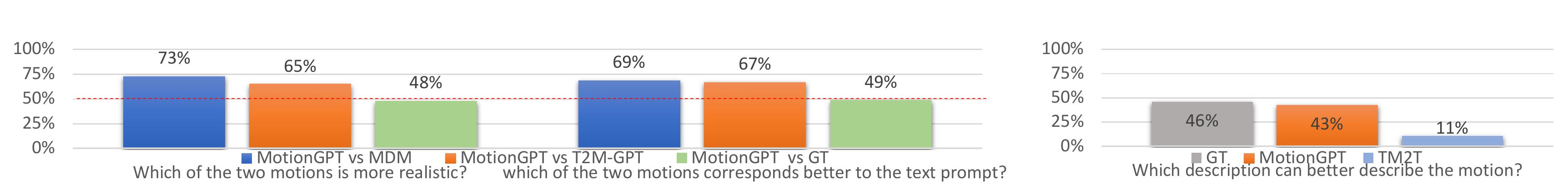}
\vspace{-10pt}
\caption{User Study. We investigate our motion quality and the alignment with test descriptions. The left part is the user study for text-to-motion. The right part is for motion captioning.}
\label{fig:appendix:user}
\end{figure}

For the comparisons of text-to-motion task, we use the force-choice paradigm to ask ``Which of the two motions is more realistic?'' and ``which of the two motions corresponds better to the text prompt?''. The provided motions are generated from 30 text descriptions from the test set of HumanML3D~\cite{Guo_2022_CVPR_humanml3d} dataset. For the comparisons of motion-to-text task, we ask 15 users to choose the motion descriptions from GT, TM2T~\cite{chuan2022tm2t}, and our MotionGPT. The motions are from the test set of HumanML3D~\cite{Guo_2022_CVPR_humanml3d} dataset.
As shown in \cref{fig:appendix:user}, in both two tasks, our MotionGPT was preferred over the other state-of-the-art methods and even competitive with the ground truth.
%

\section{Evaluation Protocols on the Uniform Motion-Language Generation.}
\label{sec:appendix:data: details}
\begin{figure}[h]
    \centering
    \includegraphics[width=\linewidth]{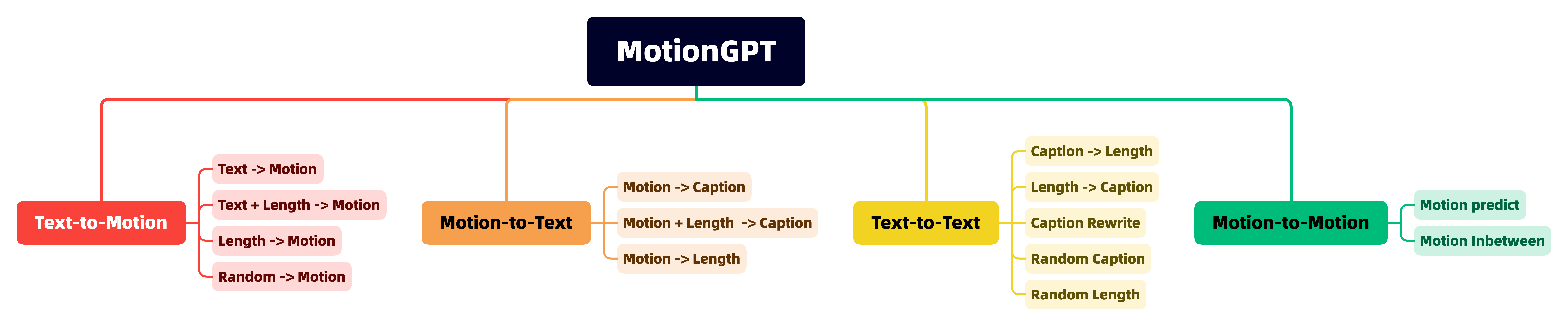}
    \caption{Protocols for multiple motion-language tasks. For each task, we follow \cref{tab:appendix:prompts} to process the previous datasets~\cite{AMASS_ICCV2019, Guo_2022_CVPR_humanml3d} into the instruction-based data.}
    \label{fig:appendix:dataset}
\end{figure}

We propose a protocol to evaluate our unified MotionGPT on multiple motion-language generation tasks. Upon previous datasets~\cite{Guo_2022_CVPR_humanml3d,Plappert2016kit,AMASS_ICCV2019}, we build an instruction motion-language dataset, which is composed of 14 core tasks (\cref{fig:appendix:dataset}) for now. As shown in \cref{tab:appendix:prompts}, each core task has dozens of instruction prompts (\cref{tab:appendix:prompts}). We will release the pre-processed dataset.

\begin{table}[H]
\centering
\resizebox{\columnwidth}{!}{%
\begin{tabular}{@{}lll@{}}
\toprule
Task & Input & Output
\\ \midrule
\multirow{3}{*}{Text-to-Motion}& Give me a motion that corresponds to [caption]. &\multirow{3}{*}{[motion] }
\\ & Demonstrate a sequence of movements that depict [caption]. &
\\ & I need a human motion that conveys [caption]. Can you generate it for me?&
\\ \midrule \multirow{2}{*}{Text-to-Motion w/ length}& Give me a motion that lasts for approximately [frames] frames. The caption is: [caption]. &\multirow{2}{*}{[motion] }
\\ & Please create a motion that lasts [seconds] seconds and illustrates [caption]. &
\\ \midrule \multirow{2}{*}{Length-to-Motion }& Show me a motion that lasts for no more than [frames] frames. &\multirow{2}{*}{[motion] }
\\ & Create a motion that has a duration of [seconds] seconds. &
\\ \midrule \multirow{2}{*}{Radnom Motion }& Give me motions as you like. &\multirow{2}{*}{[motion] }
\\ & Produce actions that are not prescribed. &
\\ \midrule \multirow{2}{*}{Motion-to-Text}& Give me a summary of the motion being displayed in [motion] using words. &\multirow{2}{*}{[caption] }
\\ & Describe the motion illustrated in [motion] in natural language. & 
\\ \midrule \multirow{2}{*}{Motion-to-Text w/ length}& Describe the movement portrayed in [motion]that lasts [frames] frames. &\multirow{2}{*}{[caption] }
\\ & What is happening in [motion] for a length of [seconds] seconds? &
\\ \midrule \multirow{2}{*}{Motion-to-Length}& What is the duration of [motion]'s gestures in frames? & There are [frames] frames in the motion.
\\ & What is the total duration of  [motion]'s body movements in seconds? & The motion lasts for [seconds] seconds.
\\ \midrule \multirow{2}{*}{Caption-to-Length}&How many frames are expected for the motion that matches [caption]?& The duration is estimated to be around [frames] frames.
\\ & Given [caption], provide the anticipated second duration for the corresponding motion. & The motion has a length of [seconds] seconds.
\\ \midrule \multirow{2}{*}{Length-to-Caption}&What are some possible physical gestures that could be made in [frames] frames?&\multirow{2}{*}{[caption] }
\\ & What motion could be performed in [seconds] seconds? &
\\ \midrule \multirow{2}{*}{Random Caption}&Depict a motion as like you have seen it.&\multirow{2}{*}{[caption] }
\\ & Describe the motion of someone randomly.&
\\ \bottomrule
\end{tabular}%
}
\vspace{10pt}
\caption{Some examples of prompt templates in our uniform evaluation protocols.}
\label{tab:appendix:prompts}
\end{table}

\label{appedix:metrics:details}

\textbf{Metric Definitions:} We provide more details of evaluation metrics as follows. Our evaluation metrics can roughly divide to five classes including text-motion matching, generation diversity, linguistic quality, motion quality, and time cost. For the first two classes, \cite{chen2023mld} has already claims clearly and for the linguistic metrics including BLUE~\cite{papineni2002bleu}, Rouge~\cite{lin2004rouge}, Cider~\cite{vedantam2015cider}, and BertScore~\cite{zhang2019bertscore}, you can refer to their own papers for details. Here we focus on the explaination of the rest metrics.


\textbf{Motion Quality}. 
FID, MPJPE, PAMPJPE~\cite{gower1975generalized}, ACCL have been clearly explained in \cite{chen2023mld}. Thus here we focus on the  Average Displacement Error (ADE) and Final Displacement Error (FDE) refaccuracy of the predicted motion. Following previous motion prediction work\cite{yuan2020dlow, zhang2021we, ma2022multi}, ADE is defined as average L2 distance between the ground truth and predicted motion of the whole sequence and FDE is the L2 distance between the ground truth and predicted motion in the last frame.

\textbf{Time Costs}. To evaluate the computing efficiency of our models, especially the inference efficiency, we calculate average Frames Per Second (FPS) when generating motions.  In our case, we calculate FPS on the test set of HumanML3D~\cite{Guo_2022_CVPR_humanml3d}, set the batch size to one, and ignore the time cost for model and dataset loading parts.

\section{Details on MotionGPT Models}
\label{sec:appendix:model:details}

\subsection{Implementation Details}
Besides the MotionGPt with 220M parameters, we implement a smaller model that reduces the model dimension with $d_{\text{model}} = 512$, $d_\text{ff} = 2048$ with only 6 layers in encoder and decoder, as well as a larger model with 770 million parameters, which increases the model dimensions with $d_\text{model} = 1024$, $d_\text{ff} = 4096$, $d_\text{kv} = 64$, 24 layers for each transformer. Except for the training iterations during the instruction tuning stage, the other settings are the same. Please refer to \cref{tab:appendix:details} for more details.

\begin{table}[H]
\centering
\resizebox{0.65\columnwidth}{!}{%
\begin{tabular}{@{}lccc@{}}
\toprule
MotionGPT &  Small & Base & Large
\\ \toprule
Backbone & Flan-T5-Small & Flan-T5-Base & Flan-T5-Large \\
Training Batch Size & 64 &16 &4 
\\
Model Size& 60M&220M&770M
\\
Pre-training - Iterations & 300K& 300K& 300K
\\
Pre-training - Learning Rate & 2e-4& 2e-4& 2e-4
\\
Instruction Tuning - Iterations & 200K& 300K& 400K
\\
Instruction Tuning - Learning Rate & 1e-4& 1e-4& 1e-4
\\
\midrule
Motion Vocabulary Number $V_m$ & 512 & 512 & 512 \\
Motion Codebook Dimension & 512 & 512 & 512\\
\bottomrule
\end{tabular}%
}
\vspace{10pt}
\caption{Hyperparameters for different MotionGPTs. We train these models on 64 Tesla V100 GPUs. The smaller model size lowers the computational requirements and thus provides faster inference ($cf.$ \cref{sec:inferencetime}). According to Tab. 6, the base MotionGPT model is the best one for overall tasks. However, we believe this could be caused by the small amount of current motion datasets. The large model could achieve the best performance when the amount of data comes up to millions or even billions. }
\label{tab:appendix:details}
\end{table}

\end{document}